\documentclass{article}

\usepackage{arxiv}
\usepackage{cite}
\usepackage{amsmath,amssymb,amsfonts}
\usepackage{algorithmic}
\usepackage{graphicx}
\usepackage{textcomp}

\usepackage{bm}

\usepackage[utf8]{inputenc} 
\usepackage[T1]{fontenc}    
\usepackage{hyperref}       
\usepackage{url}            
\usepackage{booktabs}       
\usepackage{amsfonts}       
\usepackage{nicefrac}       
\usepackage{microtype}      
\usepackage{lipsum}
\usepackage{graphicx}
\graphicspath{ {./images/} }

\usepackage{multirow}
\usepackage{booktabs}
\usepackage{url}
\usepackage{longtable}
\usepackage{pdflscape}
\usepackage{array}
\usepackage{subfig}
\usepackage{amsmath}

\usepackage{cite}
\usepackage{amsmath,amssymb,amsfonts}
\usepackage{algorithmic}
\usepackage{graphicx}
\usepackage{textcomp}
\usepackage{comment}
\usepackage[table]{xcolor}
\usepackage{makecell}

\title{Spatio-Temporal Attention Graph Neural Network: Explaining causalities with Attention}

\author{
Kosti Koistinen \\
  Aalto University School of Science \\ Computer Science Department \\ P.O.Box 11000, 00076 \\ AALTO, Finland \\
  \texttt{kosti.koistinen@aalto.fi}
   \And
 Kirsi Hellsten \\
  Aalto University School of Science \\ Computer Science Department \\ P.O.Box 11000, 00076 \\ AALTO, Finland \\
  \texttt{kirsi.hellsten@aalto.fi}
   \And
Joni Herttuainen \\
  Aalto University School of Science \\ Computer Science Department \\ P.O.Box 11000, 00076 \\ AALTO, Finland \\
  \texttt{joni.herttuainen@aalto.fi}
  \And
  Kimmo K. Kaski \\
  Aalto University School of Science \\ Computer Science Department \\ P.O.Box 11000, 00076 \\ AALTO, Finland \\
  \texttt{kimmo.kaski@aalto.fi}
}

\begin{document}
\maketitle
\begin{abstract}
Industrial Control Systems (ICS) underpin critical infrastructure and face growing cyber–physical threats due to the convergence of operational technology and networked environments.
While machine learning–based anomaly detection approaches in ICS shows strong theoretical performance, deployment is often limited by poor explainability, high false-positive rates, and sensitivity to evolving system behavior, i.e., baseline drifting. We propose a Spatio-Temporal Attention Graph Neural Network (STA-GNN) for unsupervised and explainable anomaly detection in ICS that models both temporal dynamics and relational structure of the system.
Sensors, controllers, and network entities are represented as nodes in a dynamically learned graph, enabling the model to capture inter-dependencies across physical processes and communication patterns.
Attention mechanisms provide influential relationships, supporting inspection of correlations and potential causal pathways behind detected events.
The approach supports multiple data modalities, including SCADA point measurements, network flow features, and payload features, and thus enables unified cyber–physical analysis.
To address operational requirements, we incorporate a conformal prediction strategy to control false alarm rates and monitor performance degradation under drifting of the environment.
Our findings highlight the possibilities and limitations of model evaluation and common pitfalls in anomaly detection in ICS.
Our findings emphasise the importance of explainable, drift-aware evaluation for reliable deployment of learning-based security monitoring systems.
\end{abstract}


\section{Introduction}
Modern societies rely on uninterrupted functioning of interconnected critical infrastructure, such as 
electric power grids, water treatment plants, and manufacturing systems \cite{GallaisFiliol2017_CriticalInfrastructure}. A disruption in these Operational Technology (OT) systems can cascade into severe economic, social, and physical consequences, from prolonged power outages to contaminated water supplies. Over the past decade, cyberattacks such as Stuxnet \cite{Langner2011Stuxnet}, Industroyer \cite{industroyer} and the Colonial Pipeline incident \cite{Bellamkonda2024Colonial} have demonstrated that threats once limited to Information Technology (IT) networks can now directly impact the physical world, such as equipment damage or even threats to human life \cite{macro}. During the past decade, cyberattacks on OT networks have been reported to have increased five fold from 300 annually to 1600 \cite{drago2024b}. The actual 
scale is likely to be significantly higher, as many OT intrusions remain unreported or undiscovered due to limited monitoring capabilities. By 2024, operational disruption had become routine: 50–75\% of ransomware incidents caused partial shutdowns, and approximately 25\% resulted in complete production stoppages, causing significant financial damage \cite{drago2024b}.

Industrial Control Systems (ICS) form the technological backbone of critical infrastructure and are often the target of 
cyberattacks against OT systems. They regulate physical processes through sensors, actuators, and Programmable Logic Controllers (PLCs) and are maintained through Supervisory Control and Data Acquisition (SCADA). Traditionally isolated from external networks, the OT systems once enjoyed a degree of "security by separation." However, the shift toward networked automation, remote management, and the Industrial Internet of Things (IIoT) has converged IT and OT networks, allowing adversaries to move laterally from corporate systems to industrial environments. This development has exposed ICS environments to a wide spectrum of cyber threats \cite{Selim2021Anomaly}. In addition, OT environments often rely on legacy hardware, strict safety protocols, and systems that cannot be easily updated or patched, which further increases their vulnerability to cyberattacks. 

ICS threats range from malware infections and ransomware to unauthorised remote access, data manipulation, and process disruption. In many cases, attackers exploit vulnerabilities in outdated software, weak authentication, or insecure network configurations that were never designed with cybersecurity in mind. Typical weaknesses include 
outdated protocols that allow unintended access or manipulation of control traffic. The types of attacks are commonly divided 
into network-based and physical-based attacks \cite{macro}. The former includes Denial of Service (DoS), injection, and Man-in-the-Middle attacks, while the latter include stealth attacks, data tampering, and damage attacks.

To detect and mitigate these complex attacks, Intrusion Detection Systems (IDS) are widely used in modern industrial cybersecurity. An IDS typically consists of the monitoring, pre-processing, and detection phases \cite{towards}. Among various detection approaches, such as signature-based, rule-based, and hybrid-based, anomaly-detection-based IDS have gained significant attention for their ability to learn normal operational behavior and discover deviations that may signal attacks, intrusions, or malfunctions \cite{Pinto2023Survey}. In OT networks, this capability is crucial, as anomalies are often subtle irregularities rather than clear malicious signatures. They may appear as small fluctuations in sensor data, unexpected timing patterns, unusual command sequences, or deviations in process variables that remain within protocol limits but still indicate unsafe or suspicious behavior.

There are various approaches that 
have been applied to detect and prevent cyber intrusions, including 
statistical modeling, Bayesian inference, and rule-based systems. These methods often rely on predefined assumptions about normal and abnormal behavior \cite{hu2018icsids}. However, as 
industrial systems become 
more complex and dynamic, such fixed models struggle to capture the nonlinear and time-varying nature of real-world operations \cite{ding2018icps}. In contrast, machine learning–based approaches have attracted widespread interest for their ability to automatically learn patterns from data and adapt to evolving system behavior \cite{Pinto2023Survey}. These methods can uncover correlations between multiple variables, which makes them particularly suitable for anomaly detection in ICS.

Traditional machine learning approaches include, for example, k-nearest neighbors, Random Forests, and Support Vector Machines \cite{classicalmethods}. However, despite their efficiency in classification, these methods are insufficient to model the temporal dependencies that are inherent in OT traffic. They are also sensitive to imbalanced datasets such that a 
new unseen anomaly often remains undetected. Furthermore, in most OT environments, the majority of traffic is benign, while only a small fraction represents attacks. This imbalance can lead to biased classifiers that fail to detect rare but critical anomalies. 

To overcome these limitations, Deep learning approach has 
emerged as a promising solution. 
Autoregressive architectures such as Long Short-Term Memory (LSTM) networks \cite{lstmnetworks} and other Recurrent Neural Network approaches (RNNs) \cite{otherrnn} can 
capture complex temporal patterns among variables, which allows the model to understand how changes in one part of the system influence the rest. However, the performance of these models also suffers from unbalanced data. 
Some other methods include autoencoders, Generative Adversarial Networks, and a mixture of these models, but all have similar limitations. More recently, transformer-based architectures have become popular because of their ability to model long-range dependencies using self-attention mechanisms. Their success in natural language processing has motivated research into their application for anomaly detection in time-series and network data, where sequential dependencies and contextual relationships are crucial. Transformer models, and in particular adaptations of language models, show the potential to capture complex semantic patterns in network traffic representations \cite{KHEDDAR2025103347}.

Beyond and within sequential approaches, graph-based deep learning provides a fundamentally different way to represent and analyze OT systems. By modeling the system as a graph, where nodes represent entities (such as devices or sensors) and edges represent their relationships or communications, a more realistic and structured view of the environment can be obtained \cite{ObjectDetection, xing2026human}. Graph-based models are able to uncover non-linear correlations and long-range dependencies that traditional time-series or tabular approaches often miss. Graph Neural Networks (GNNs), such as Graph Convolutional Networks (GCNs) \cite{GCNpaperi} and Graph Attention Networks (GATs)\cite{veličković2018GAT}, exploit this representation by learning how information propagates through the network structure. GCNs aggregate neighborhood information to capture local dependencies, while GATs extend this approach by applying attention mechanisms to weigh the importance of different connections dynamically. Through these formulations, GNNs can effectively model both the topology and interactions within the system, and thus enable a more accurate detection of anomalous behavior that may emerge across multiple interconnected entities. For a more detailed
review of current methods, see \cite{LamichhaneEberle2024}.

Although harnessed with state-of-the-art machine learning, there are still some issues to consider before their usage for anomaly detection in ICS, 
such as the lack of open high-quality datasets for research and high false alarm rates (for other challenges, see \cite{Quincozes2025_CPS_Datasets_Taxonomy}). In addition, deep learning introduces challenges of its own, related to its explainability and interpretability. As models become more complex and rely on multilayer architectures, their internal decision-making processes become opaque. In ICS environments, operators must understand why an alert was triggered, and this lack of transparency creates a significant barrier to adoption. Explainability methods aim to improve the trustworthiness, interpretability, and accountability of machine learning models by providing human-understandable insights into how they reach their conclusions \cite{Linardatos2021_ExplainableAI}.

However, the application of explainability techniques into ICS remains a challenge. First, ICS traffic is often high-dimensional and highly contextual, making it difficult to map model outputs to meaningful operational features. Second, many explainability tools are computationally expensive or unstable when applied to time-series or graph-based deep learning models. Third, explanations must be not only technically accurate but also domain-relevant, i.e., 
operators need actionable insights, not abstract attributions. As a result, despite significant progress, current explainability solutions often do not meet the stringent requirements of industrial environments. More research is needed to develop lightweight, reliable, and domain-aware explainability mechanisms that can support real-time decision-making and foster operator trust in AI-driven anomaly detection.

To address the aforementioned challenges, we propose an unsupervised GNN-based framework that uses graph-oriented machine learning for explainable anomaly detection in ICS. The model constructs a graph representation of the system that enables learning of the relationships among sensors, actuators, and process variables. Within this architecture, attention mechanisms are employed to extract the most influential dependencies in the graph, allowing the model to focus on critical interactions during anomaly scoring. By examining the resulting correlation structures, we analyse how the model’s learned relationships align with known causal dependencies in the industrial process. This facilitates transparent system-level anomaly detection, which traditional models might overlook. Furthermore, the framework is tunable as it can operate on SCADA-point data for point-level anomaly detection, on netflow data for passive network monitoring, or on both modalities simultaneously through a multimodal configuration.

This study is organised such that in  Section~\ref{sec:RelatedWork} we discuss related works in the field of Explainable AI. Next, in Section~\ref{sec:Methodology} we introduce our model architechture and evaluation strategy. 
In Section~\ref{sec:Benchmark_Data} we present the data we use for benchmarking the model, 
followed by presenting the results and analysis of the acquired graph representations 
in Section~\ref{sec:Results}. Then in Section~\ref{sec:Discussions} we discuss the methodological and practical issues encountered during the analysis and reflect more broadly the common issues in Machine Learning anomaly detection. 
In Section~\ref{sec:Conclusions_FutureWork} we draw conclusions and on what could be the focus of future work.

\section{Related Work}
\label{sec:RelatedWork}

In this section, we provide a short review of the most relevant work on explainable artificial intelligence (XAI). 
Although the literature on XAI is extensive, (see 
e.g., \cite{XAI1, XAI3}), 
only recently have cybersecurity and IDS applications begun to receive dedicated attention. Here, our aim is to highlight the works that explore explainability, specifically, for non-experts and experts in IDS and of OT environments.

Explainable AI as a field emerged formally in 2004 \cite{van_lent_explainable_2004}, but its development accelerated significantly in the last decade alongside the rise of deep learning. The "black box"  nature of the deep learning models grew an interest for trustworthy and explainable AI in various fields, e.g., in medical sciences, finance and autonomous systems \cite{arrieta2020xai}. A widely accepted taxonomy categorises XAI methods into intrinsic (\textit{ante-hoc}) and \textit{post-hoc} models \cite{XAI4}. Intrinsic explanations arise directly from the model architecture through weights, rule structures, or built-in interpretability constraints. The model itself is designed to be transparent. For example, these models include classifiers and regressors. In contrast, \textit{post-hoc} approaches aim to explain the model's outputs via various tactics. Early contributions include game-theory approaches, in which SHAP explanations are the most popular for explaining the importance of features. Another popular type of \textit{post-hoc} -approaches includes gradient- and decomposition-based techniques, where backpropagated gradients are modified or analysed to attribute importance \cite{Guo2018LEMNA}. Other examples include perturbation-based explanations
\cite{marino2018adv,DEEPAID}. The latter
raises an important point 
that most of the XAI-methods are for supervised setting, while 
in most of the real-world ICS systems labeled data are unrealistic assumptions. The authors provided an unsupervised fine-tuning module that could be used in problematic features, allowing for model adjustment without exhaustive re-training.

The XAI literature for graph deep learning includes several \textit{post-hoc} explanation techniques designed to interpret the predictions of GNNs. Many of these approaches rely on graph masking, where the goal is to learn masks on 
edges, nodes, or features to identify the substructures most influential in 
a model’s decision. One of the most widely cited methods is GNNExplainer \cite{2019lesko}, a model-agnostic explainer that is applicable to any GNN architecture. By optimizing soft masks on 
edges and features, GNNExplainer extracts subgraph-level explanations that find 
key structural components and node attributes driving the output of the model. 
This method has been adopted in cybersecurity contexts, including IDS, as demonstrated in \cite{Yu2025GNNexPIDS}. A related method is PG-Explainer \cite{luo2020PG}, which differs from GNNExplainer by training a parametric explanation network that generalises across instances rather than optimizing a mask separately for each prediction. The optimization strategy improves scalability and stability while retaining the ability to identify influential edges. PG-Explainer has been utilised in IDS research, for example, in \cite{Kaya_2024_PGIDS}.

In OT environments, the application of graph explainers is much more limited. A notable exception is KEC (K-hop Explanation with Convolutional Core) \cite{wang2022faithful}, which was applied to anomaly detection in the SWaT industrial control benchmark dataset \cite{mathur2016swat}. Unlike the masking-based paradigm, KEC constructs a surrogate linear model that approximates the local behavior of the GNN and derives explanations through gradient-based attribution. The authors introduce a formal notion of faithfulness, a measure of how well an explainer preserves model behavior and 
show that KEC achieves higher faithfulness than existing explainers.

A common challenge among GNN explanation methods is that many of them provide partial explanations, focusing only on one dimension—edges, nodes, or features—without offering a unified view. The ILLUMINATI framework \cite{Ill} addresses this limitation by producing comprehensive explanations that consider the contribution of node importance, edge importance, and node-attribute 
together. Designed specifically for cybersecurity use cases, it extends traditional masking approaches with a richer explanatory structure. Comparative evaluations of GNN explainers for IDS have also emerged. For example, recent work in \cite{Galli2025} finds that GraphMask \cite{2020graphmask} performs particularly well for DoS and DDoS attack detection, outperforming other explainers in robustness and interpretability. However, despite this promising result, we did not find substantial evidence of GraphMask being applied more broadly in IDS or OT-focused literature.

Finally, a branch of graph deep learning approaches uses attention mechanisms \cite{vaswani2017attention} as a tool for generating explanations. Attention mechanism allows for a model to assign different importance weights to different nodes or edges, highlighting which relationships it considers most relevant during prediction. Graph Attention Networks (GATs) are build on this idea by using attention to reveal correlations between learned embeddings \cite{veličković2018GAT}. Some models, including the Graph Deviation Network (GDN) \cite{deng2021graph}, which also inspires the present study, 
apply attention mechanisms to time series for 
identifying variable-level dependencies and highlighting anomalous patterns. This approach captures localised deviations in sensor behavior
using both structural relationships and temporal dynamics within OT systems. A very recent approach, PCGAT \cite{act14050210}, extends attention-based reasoning by modeling ICS through multi-level physical-process and controller graphs, to enable both anomaly detection and anomaly localization via attention patterns. The authors highlight several limitations of typical attention-based methods. They argue that attention weights learned purely from data do not necessarily correspond to the true causal or physical relationships in ICS, and therefore may produce explanations that are misleading from an operational perspective. This can create difficulties in identifying the actual sources of anomalies and understanding how they propagate through the system. Furthermore, they claim that many existing GAT-based anomaly detectors rely on unrealistic fully connected sensor graphs, resulting in high computational cost, redundancy, and limited interpretability. These models also fail to incorporate the hierarchical and process-driven structure of ICS, reducing their reliability and diminishing the usefulness of attention weights as explanations. In short, the complexity of graph-based deep learning introduces several challenges, which our study seeks to address.
%
%
%
%
%
%
%
%
%
%
%
%
%
%
%
%
%
%
%
\section{Methodology}
\label{sec:Methodology}
\begin{figure*}[ht]
    \centering
    \includegraphics[width=0.90\linewidth]{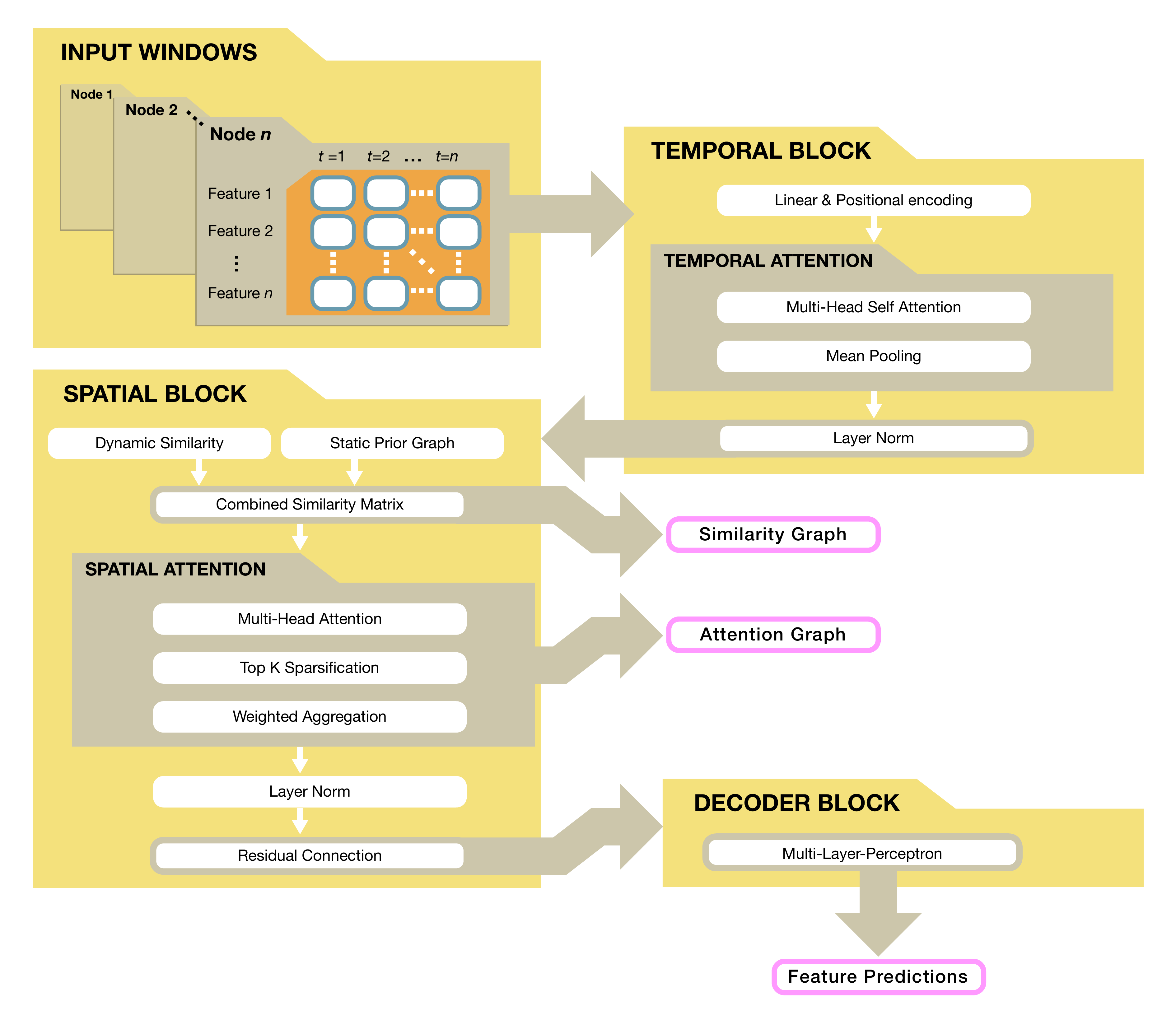}
    \caption{A schematic overview of the STA-GNN model architecture. The workflow illustrates the processing stages from input windows to the decoder producing predictions. The intermediate blocks employ a two-phase attention mechanism that generates two complementary graphs, enabling inspection of the model's decision making.}
    \label{fig:palapala}
\end{figure*}

Here we propose the 
Spatio-Temporal Attention Graph Neural Network (STA-GNN),
designed to capture both temporal dependencies and dynamic spatial correlations among sensors or devices (henceforth entities) in multivariate time series data. The STA-GNN is inspired by the Graph Deviation Network\cite{deng2021graph}
and Graph Attention Network\cite{veličković2018GAT}, with several modifications combining 
temporal attention mechanisms with an adaptive graph construction strategy that learns context-dependent relationships between entities. In this section, we will explain the model architecture and anomaly detection methodology, and the model framework is illustrated in Fig.~\ref{fig:palapala}.

\subsection{Architecture}
Each node in the graph corresponds to an entity and is associated with a multivariate feature vector. Unlike GDN, our model allows for multiple features per entity, which in turn allows 
for multi-modality. Specifically, at each timestep $t$, an entity $i$ is represented by a feature vector $x_{t,i} \in \mathbb{R}^{F}$, where $F$ denotes the number of observed variables for that $i$ (e.g. continuous measurements and Boolean indicators). Over a sliding window of length $W$, the input tensor therefore takes the form $X \in \mathbb{R}^{B \times W \times N \times F}$, where $N$ is the number of $i$ and $B$ is the batch size. Throughout the model, the nodes are treated as feature-bearing entities whose representations are progressively transformed into latent embeddings that jointly encode temporal dynamics and inter-dependencies. The model first applies a linear projection $H$ at each timestep~$t$:
%
\begin{align}
    H_t = \mathrm{Linear}(X_t) + P_t,
\end{align}
where $P_t$ represents a learnable positional embedding for the timestep $t \in \{1,\dots,W\}$ that encodes the temporal order within the observation window. Next, we go through in detail the stages of the anomaly detection process from the input window to the temporal, spatial, and decoder blocks of the STA-GNN model architecture.
\subsubsection*{Temporal Block}
To model temporal dependencies, each nodes' time series within the observation window is processed by a multi-head self-attention mechanism (MHA), inspired by the Transformer architecture and originally developed for natural language processing \cite{vaswani2017attention}. MHA enables each timestep in a node’s sequence to attend to every other (past) timestep within the window, allowing the model to capture both short-term fluctuations and long-range temporal dependencies without relying on recurrence. In practice, we apply causal masking in the temporal attention module so that a timestep cannot attend to future observations, preventing information leakage or data snooping.

Formally, given the linear projection $H_t$ for a single entity, the attention module constructs representations for query $Q$, key $K$ and value $V$ 
through learned linear projections:
\begin{align}
Q_t &= W_Q H_t, \quad K_t = W_K H_t, \quad V_t = W_V H_t,
\end{align}
where $W_Q, W_K, W_V \in \mathbb{R}^{d\times d}$ are learnable parameter matrices, and $d$ denotes the embedding dimension of the latent representation. The linear projection operates across the feature dimension $F$. The attention weights are computed as scaled dot-products between queries and keys:
\begin{align}
\alpha_t = \mathrm{softmax}\left(\frac{Q_t K_t^\top}{\sqrt{d}}\right),
\end{align}
which measure the degree of relevance between every timestep pairs. 
These weights are then used to form a weighted sum of the value vectors:
\begin{align}
H'_t = \alpha_t V_t,
\end{align}
producing an updated temporal representation where each timestep encodes information aggregated from all others. To capture seasonal, weekly, and daily fluctuations, multiple attention heads are used in parallel, each operating on a different subspace of the embedding dimension. The outputs of these heads are linearly combined:
\begin{align}
H' = \mathrm{MHA}(Q_t, K_t, V_t) = \mathrm{Concat}(\text{head}_1, \dots, \text{head}_h)\, W_O,
\end{align}
where $W_O \in \mathbb{R}^{(h\cdot d_h)\times d}$ projects the concatenated result back to the model dimension. The resulting representation is then aggregated across timesteps (e.g., via mean pooling) and normalised through a Layer Normalization (LN) operation, yielding the final temporally encoded features
\begin{align}
H = \mathrm{LN}\left(\frac{1}{W}\sum_{t=1}^{W} H'[t]\right),
\end{align}
where $H \in \mathbb{R}^{B\times N\times d}$ represents the temporally contextualised embedding for each entity. This tensor $H$ is the output of the temporal feature extractor and serves as the input to the subsequent spatial attention stage, which models the inter-entity dependencies.
\subsubsection*{Spatial Block}
Unlike conventional GNNs that rely on static graphs, the STA-GNN constructs dynamic spatial graphs based on both the contextual similarity $S_\mathrm{ctx}$ and static similarity $S_\mathrm{st}$. For each sample $b$, the dynamic contextual similarity is computed from the temporally encoded features as
\begin{align}
    S_\mathrm{ctx}^{(b)} = H_b H_b^\top,
\end{align}
where $H_b \in \mathbb{R}^{N \times d}$ denotes the slice of $H$ corresponding to the batch element $b$. In addition to the
dynamic similarity, the model also supports an optional external static prior graph $A_{\mathrm{static}} \in \mathbb{R}^{N \times N}$, which 
can encode domain knowledge about the entity connectivity, physical topology, or known relationships. When provided, the entries of $A_{\mathrm{static}}$ are normalised and incorporated directly as the static similarity term. If no external graph is supplied, the model instead learns a static entity embedding matrix $E \in \mathbb{R}^{N \times d}$, from which a static similarity is constructed as 
\begin{align}
    S_\mathrm{st} = E E^\top,
\end{align}
which corresponds to a (scaled) cosine similarity after $\ell_2$-normalization of the rows of the embedding matrix $E$. If prior is introduced, $S_{\mathrm{st}}$ is passed with normalised values of $A_{\mathrm{static}}$. The combined similarity matrix is then given by
\begin{align}
    S^{(b)} = S_\mathrm{ctx}^{(b)} + \lambda S_\mathrm{st},
\end{align}
where $\lambda$ is a learnable scaling parameter. The model thus learns to adaptively balance between dynamic contextual dependencies and static structural patterns.

To propagate information among entities, the model applies another attention mechanism over the temporally encoded representations $H$ to capture spatial dependencies. In this phase, the queries, keys, and values are newly projected from $H$ using distinct learnable matrices $\left(W_Q^{\mathrm{(sp)}},W_K^{\mathrm{(sp)}}, W_V^{\mathrm{(sp)}}\right)$, which allow 
each entity to attend to all others based on their recent temporal behavior. We employ multi-head scaled dot-product attention over entities (rather than a GAT-style additive attention with a LeakyReLU nonlinearity). Concretely, for each head, queries, keys, and values are obtained as
\begin{align}
Q_\mathrm{sp} &= W_Q^{\mathrm{(sp)}} H, \quad
K_\mathrm{sp} = W_K^{\mathrm{(sp)}} H, \quad
V = W_V^{\mathrm{(sp)}} H,
\end{align}
and the attention logits are computed via scaled dot-products between entities. The resulting attention scores are modulated by the similarity prior $S^{(b)}$, yielding the dynamic attention matrix
\begin{align}
A^{(b)} = \mathrm{softmax}\left(\frac{Q_\mathrm{sp}^{(b)} K_\mathrm{sp}^{(b)\top}}{\sqrt{d}\, T} + S^{(b)}\right),
\end{align}
where $T$ is a learnable temperature parameter controlling the sharpness of attention. 
To enhance sparsity and interpretability, only the top-$k$ most relevant neighbors (i.e., with the highest attention weights) are kept for each node, ensuring efficient message passing and reducing noise from weak connections. For multi-head attention, this procedure is applied independently per head; the resulting attention weights can be averaged across heads for interpretability.

The spatially constructed features for each sample $b$ and entity $i$ are then given by
\begin{align}
    H^{\mathrm{(sp)}}_{b,i,:} = \sum_{j=1}^N A^{(b)}_{i,j} V_{b,j,:} + \beta H_{b,i,:},
\end{align}
or, in matrix form,
\begin{align}
    H^{\mathrm{(sp)}} = AV + \beta H,
\end{align}
where $\beta$ is a residual weighting factor. Thus, each $H^{(\mathrm{sp})}_{b,i,:}$ is a learned spatio-temporal feature vector for entity $i$, obtained as an attention-weighted aggregation of its neighbors’ value embeddings plus a residual contribution from its own temporal representation. The resulting tensor $H^{(\mathrm{sp})} \in \mathbb{R}^{B\times N \times d}$ encodes both temporal and spatial dependencies for each entity.
\subsubsection*{Decoder Block}
Finally, the normalised representations are passed through a fully connected multilayer perceptron (MLP) decoder applied independently to each entity. For each sample $b$ and entity $i$, we compute
\begin{align}
\hat{y}_{b,i} = f_\theta\!\left(H^{(\mathrm{sp})}_{b,i,:}\right),
\end{align}
where $f_\theta$ denotes a two-layer feed-forward network with nonlinearity (ReLU) between layers. In matrix form, this can be written as
\begin{align}
\hat{Y} = \mathrm{MLP}(H^{(\mathrm{sp})}) \in \mathbb{R}^{B\times N\times F},
\end{align}
yielding one output per node feature and sample based on the final spatio-temporal feature representation.
\subsection{Training Objective}
Each entity $i$ may contain both continuous and Boolean features, and the loss aggregates reconstruction errors across these feature dimensions. This design allows heterogeneous variables to contribute appropriately to the training signal while preserving a unified node-level representation in the graph. For example, exogeneous temporal features may be appended to node features.

The model is trained in a semi-supervised setting, using only data assumed to represent normal system behaviour. The learning objective is to minimise the difference between the model’s reconstructed feature values
$\hat{Y}_b$ and the observed values $Y_b$ for each batch element $b \in \{1, \dots, B\}$. Because the dataset may include both continuous-valued features and
Boolean/indicator features, we employ a composite loss function,
\emph{MixedLoss}, which combines a mean–squared error (MSE) term for
continuous features and a binary cross–entropy (BCE) term for Boolean
features. Let $\mathcal{C}$ denote the indices of continuous features and
$\mathcal{B}$ the indices of Boolean features.
The training loss for a single window is
\begin{equation}
\begin{aligned}
\mathcal{L}_{\text{mixed}}
    &= \gamma_{\text{cont}} \cdot \frac{1}{|\mathcal{C}|}
       \sum_{(i,f)\in \mathcal{C}} (\hat{Y}_{b,i,f} - Y_{b,i,f})^2 \\
    &\quad + \gamma_{\text{bool}} \cdot \frac{1}{|\mathcal{B}|}
       \sum_{(i,f) \in \mathcal{B}}
       \mathrm{BCE}(\hat{Y}_{b,i,f}, Y_{b,i,f})
\end{aligned}
\end{equation}
where $\gamma_{\text{cont}}$ and $\gamma_{\text{bool}}$ weight the relative influence of
continuous and Boolean feature types.
MixedLoss ensures that each feature type contributes appropriately to the
learning signal. At inference time, we use the same \emph{MixedLoss} formulation both for the
scalar anomaly score and for per-entity explanations, ensuring that the
detection objective is aligned with the training objective.

\subsection{Anomaly Scoring}

For each sliding window~$w$, we compute feature-wise reconstruction
errors and aggregate them into a per-entity MixedLoss contribution. For a
continuous feature $f \in \mathcal{C}$ of entity $i$, the reconstruction error
is defined as
\[
e_{w,i,f} = (\hat{Y}_{w,i,f} - Y_{w,i,f})^2,
\]
and for a Boolean feature $f \in \mathcal{B}$ of entity $i$, we define
\[
e_{w,i,f} = \mathrm{BCE}(\hat{Y}_{w,i,f}, Y_{w,i,f}).
\]
Each $e_{w,i,f} \ge 0$ therefore represents the MixedLoss error contribution of
feature $f$ of entity~$i$ for window~$w$.

The per-entity reconstruction error is obtained by aggregating feature-wise
errors using the same weighting scheme as in training:
\[
e_{w,i}
=
\gamma_{\text{cont}} \cdot \frac{1}{|\mathcal{C}_i|}
\sum_{f \in \mathcal{C}_i} e_{w,i,f}
\;+\;
\gamma_{\text{bool}} \cdot \frac{1}{|\mathcal{B}_i|}
\sum_{f \in \mathcal{B}_i} e_{w,i,f},
\]
where $\mathcal{C}_i$ and $\mathcal{B}_i$ denote the sets of continuous and
Boolean features associated with entity $i$, respectively. The model can
therefore be used either by aggregating the errors per node or by detecting
anomalies at the node--feature level. An overall anomaly score for the window is finally obtained by averaging the
per-entity losses:
\[
s_w = \frac{1}{N} \sum_{i=1}^N e_{w,i} \ \ .
\]
Higher values of $s_w$ reflect a greater deviation from the behavior learned during
training. This scoring strategy follows the standard semi-supervised anomaly detection assumption that a model trained only on nominal operating data reconstructs normal behavior more accurately than behavior generated under abnormal or attack conditions. Under this assumption, attacks or other out-of-distribution events are expected to induce larger reconstruction errors than nominal samples.

\subsection{Graph Explanations}
During inference, the STA-GNN produces two complementary graph structures, the contextual similarity graph $G_{cs}$ and the attention graph $G_{a}$. In both representations, the nodes correspond to entities, whereas the dynamically evolving edges encode relationships between them.
The $G_{cs}$ captures relations between the learned temporal embeddings, reflecting how similar the recent temporal dynamics of different entities are within a given observation window. In contrast, the $G_a$ represents directed inter-entity dependencies, where edge weights encode the magnitude and direction of the learned correlations, that is, how information is propagated between entities in the latent space.

\begin{figure*}[ht!]
    \centering
    \includegraphics[width=0.5\linewidth]{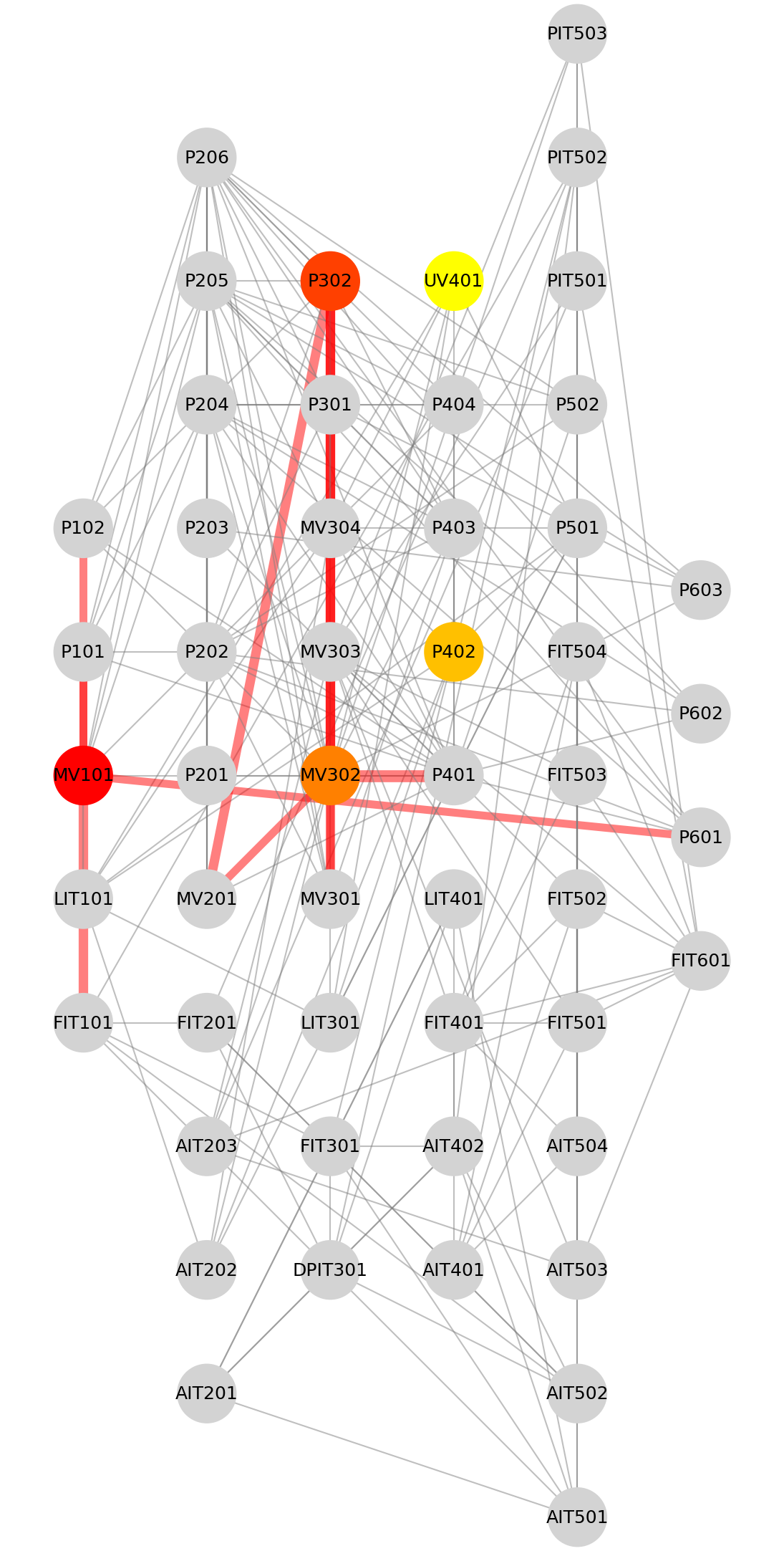}
    \caption{Example. Attack detected and contribution highest from red (highest) to yellow (lowest). The grey edges represent the learned embeddings + prior graph structure. The red edges come from the spatial attention. Only the strongest attention weights from/to anomalous nodes are plotted for interpretability. Red edge thickness reflects to strength of the attention. The graph nodes are organised and fixed by process stages in SWaT testbed dataset used in this study.}
    \label{fig:placeholder2}
\end{figure*}

Fig.~\ref{fig:placeholder2} illustrates an example of the model’s outputs during anomaly detection.
When an anomaly is detected, both graphs are visualised to highlight the underlying relational patterns. The nodes that are considered anomalous, are plotted with distinct colours, while the rest are kept at the background as grey. For interpretability, only the top five edges with highest similarity per node are retained in $G_{cs}$, ensuring sparse and readable structure. For $G_a$, the edges are filtered to include only those that originate or end at anomalous nodes. The amount of edges is restricted by topk-attention weights.

\subsection{Model Evaluation}

One of the main metrics to evaluate the performance of our model in detecting anomalies is the false positive rate (FPR) defined as follows
\begin{align}
    \mathrm{FPR}
     = \frac{\mathrm{FP}}
            {\mathrm{FP} + \mathrm{TN}},
\end{align}
where false positive (FP) is the number of incorrect predictions or false alarms, and true negative (TN) is the number of correct predictions of no alarms. In the model evaluation, the emphasis is on minimising the FPR i.e., avoiding false alarms, while still maintaining adequate anomaly detection performance. 
Furthermore, we summarise the detection quality by using the F1 score and evaluate two thresholding strategies: (i) a threshold that maximises the F1 score on validation data and (ii) a conformal-thresholding scheme based on nonconformity scores.
The F1 score combines precision and recall into a single harmonic-mean metric. Given the number of true positives (TP), false positives (FP) and false negatives (FN), the F1 score is defined as
\begin{align}
    \mathrm{F1}
    &= \frac{2 \,\mathrm{precision} \cdot \mathrm{recall}}
            {\mathrm{precision} + \mathrm{recall}}
     = \frac{2 \,\mathrm{TP}}
            {2\,\mathrm{TP} + \mathrm{FP} + \mathrm{FN}},
\end{align}
where $\mathrm{precision} = \mathrm{TP} / (\mathrm{TP} + \mathrm{FP})$ and
$\mathrm{recall} = \mathrm{TP} / (\mathrm{TP} + \mathrm{FN})$. We first compute anomaly scores $s_w$ for each window $w$ and choose a threshold that maximises the F1 score on the labeled evaluation set. This provides an unsupervised operating point that balances missed anomalies and false alarms.

\subsection{Conformal Nonconformity Scoring and Thresholding}
The reconstruction error itself does not provide a formal guarantee that all normal and attack samples are separable. To make the anomaly score operationally usable, we calibrate its threshold using inductive nonconformity scoring scheme \cite{indconform2002}. Let $s_1, \dots, s_T$  denote the anomaly scores on a set of calibration windows assumed to be normal. Then we 
define difference nonconformity scores $c$ as 
\begin{align}
    c_1 &= 0, \\
    c_t &= \max\bigl(0,\, s_t - s_{t-1}\bigr), \quad t = 2,\dots,T,
\end{align}
which emphasize sudden increases in the anomaly score blue while being 
less sensitive to slow shifts of values. Given a significance level $\alpha$, we then choose a threshold $q_\alpha$ as an upper quantile of the calibration scores, i.e., 
\begin{align}
    q_\alpha = \mathrm{Quantile}_{1-\alpha}(c_1,\dots,c_T),
\end{align}
at evaluation time declare window $t$ anomalous if $c_t \ge q_\alpha$. The benefit of the conformal approach is twofold: first it automatically adapts to the empirical score distribution and second, under exchangeability assumptions, it provides finite-sample guaranties that the probability of a false alarm does not exceed approximately significance level $\alpha$. The exchangeability assumption means that the calibration nonconformity scores
and the future nominal evaluation scores are assumed to be drawn from the same underlying distribution. In the present setting, this requires that the calibration period is representative of the nominal operating regime.

In our experiments, we choose a heuristic value $\alpha = 10^{-3}$, which yields a low false positive rate while still allowing the model to react to pronounced score increases. For example, with data sampled in 10-second intervals, this threshold corresponds to an expected false alarm roughly once every three hours under nominal conditions. Another advantage of the approach is that the threshold $q_\alpha$ is fixed by the distribution of the calibration scores. As a result, if the typical scoring behavior of the system starts to change and the evaluation scores consistently exceed their calibration levels, the number of threshold exceedance will gradually increase. This behavior is a clear indication of covariate drift, which signals that the model may no longer be well suited to the altered environment. Conventional performance metrics, such as F1-score or accuracy, cannot reveal such changes in the underlying data distribution. Although conformal thresholding does not prove attack separability, it provides a practical deployment mechanism for detecting threshold exceedance and monitoring the model degradation under drift. For a detailed description of the conformal prediction framework, see \cite{kato23a}.

\section{Benchmark Data}
\label{sec:Benchmark_Data}
The Secure Water Treatment (SWaT) testbed is one of the most widely used benchmarking datasets available for research on ICS security. It represents a scaled-down, fully operational water purification plant designed to reproduce the behavior, equipment interactions, and cyber-physical processes found in real facilities. The system produces approximately five gallons of treated water per minute and operates in six sequential process stages, each equipped with a range of sensors, such as level transmitters, pressure gauges, and water-quality probes, as well as actuators including pumps and motorised valves. The sensors and actuator names, and further detailed description of the environment, are provided in \cite{mathur2016swat}. In illustrative Figures \ref{fig:placeholder2}, \ref{fig:placeholder} and \ref{fig:both}, we have arranged the process stages horizontally, from left, process stage 1, to stage 6, right.

The SWaT datasets provide both process-layer (physical-level) measurements obtained from the SCADA/PLC level and detailed OT network traffic, including partial CIP protocol payloads. Communication between PLCs, sensors, actuators, and the supervisory SCADA layer is extensively logged, enabling the simultaneous analysis of physical process behavior and network activity. This multimodal perspective is crucial as previous work has shown that effective anomaly detection requires both physical measurements and communication patterns, since attacks may affect only a single modality or manifest across both \cite{Zhan2025CrossDomainICS}.

The 2015 SWaT dataset includes a long period of normal operation followed by a series of 41 controlled cyberattacks, targeting communication links and manipulating one or multiple process stages. These attacks range from stealthy modifications to aggressive actuator manipulation, making the 2015 SWaT dataset a challenging and realistic benchmark. The rest of the selected SWaT datasets used in our study are provided in Table~\ref{tab:data}.
\begin{table*}[t]
\centering
\caption{Overview of the SWaT datasets used in this study across physical and network modalities. Measurements from physical sensors and network traffic were aggregated and resampled to 10-second intervals.}
\label{tab:data}
\begin{tabular}{llccccc}
\toprule
SWaT & Modality & Nodes & \#Features & \#Instances & Duration & \#Attacks \\
Dataset &&&&&& \\
\midrule

\multirow{3}{*}{2015}
  & Physical          & 51 & 1  & $\sim$ 95\,000 & 7 d normal + 4 d attack & 41 \\
  & NetFlow           & 9  & 11 & $\sim$ 95\,000 & 7 d normal + 4 d attack & 41 \\
  & NetFlow+Payload   & 9  & 14 & $\sim$ 95\,000 & 7 d normal + 4 d attack & 41 \\
\midrule

\multirow{3}{*}{2017}
  & Physical          & 51 & 1  & $\sim$ 49\,000  & 6 d normal & 0 \\
  & NetFlow           & 9  & 11 & $\sim$ 17\,000  & 2 d normal & 0 \\
  & NetFlow+Payload   & 9  & 14 & $\sim$ 17\,000  & 2 d normal & 0 \\
\midrule

\multirow{1}{*}{2019 Jul}
  & Physical          & 51 & 1  & $\sim$ 1\,500   & 4 h attack & 6 \\
\midrule

\multirow{3}{*}{2019 Dec}
  & Physical          & 51 & 1  & $\sim$ 1\,300   & 4 h attack & 5 \\
  & NetFlow           & 9  & 11 & $\sim$ 1\,300   & 4 h attack & 5 \\
  & NetFlow+Payload   & 9  & 14 & $\sim$ 1\,300   & 4 h attack & 10 \\
\bottomrule
\end{tabular}
\end{table*}

\subsection{Data Pre-Processing \& Model Training}
\paragraph{Physical-level Data}
For the physical-level data, all continuous sensor values were treated as floating-point variables, while discrete control states (e.g., \textit{on}, \textit{off}, \textit{auto}) were one-hot encoded. Continuous features were scaled using min--max normalization, defined as
\begin{align}
    x' = \frac{x - x_{\min}}{x_{\max} - x_{\min}},
\end{align}
where $x_{\min}$ and $x_{\max}$ are minimum and maximum values from the training data, and $x$ is a value to normalise. The evaluation dataset was fitted with these normalization parameters.
In physical-level datasets, each node corresponds to a single sensor or actuator signal, and no additional node-level features were introduced.

\paragraph{Netflow Dataset}
For the Netflow data, an explicit design choice was required to define the node entities. We chose the set of IP addresses observed in the traffic as entities. More precisely, we selected PLC-, SCADA point-, and workstation IP addresses as individual nodes, based on prior system knowledge. All remaining traffic was aggregated into a single auxiliary node labeled \textit{Other IP}.

We extracted the features of the standard NetFlow protocol, including the source port, source IP, destination IP, transport protocol, and frame length. We restricted the feature set to flow-level metadata, as packet payloads are often encrypted and therefore unavailable. Moreover, flow-based representations significantly reduce computational costs compared to deep packet inspection \cite{UMER2017238}. From these base features, we derive the features per node. These include, for example, Shannon entropy, defined as
\begin{align}
H_{\text{src}} = - \sum_{i=1}^{k} p_i \log_b(p_i),
\end{align}
where $k$ denotes the number of distinct source ports observed within an aggregation window, and $p_i = \frac{n_i}{N}$ is the empirical probability of the source port $i$, with $n_i$ occurrences out of $N$ total flows. The rest of the derived features are presented in Table~\ref{tab:features}. We note that this is just an example, and other approaches for deriving features exist.

\begin{table*}[ht]
\centering
\caption{Aggregated node-level features for the NetFlow and NetFlow+Payload data models. All features are sampled 10 seconds interval.}
\label{tab:features}

\begin{tabular}{p{0.40\linewidth}|p{0.56\linewidth}}
\hline
\textbf{Feature} & \textbf{Description} \\
\hline
& \\

\multicolumn{2}{l}{\textit{NetFlow features}} \\ \hline
Rows sent / received & Number of flow records sent and received \\ \hline
Bytes sent / received & Total number of bytes sent and received from frame length \\ \hline
Source port entropy & Entropy of observed source ports \\ \hline
Protocol entropy & Entropy of observed protocols \\ \hline
\#Sources / \#Destinations & Number of distinct source and destination peers \\ \hline
& \\
\multicolumn{2}{l}{\textit{NetFlow + Payload features}} \\ \hline
CIP byte entropy & Shannon entropy of the CIP payload bytes. For example, typical message could be 10x4 bytes. \\ \hline
CIP value mean & Mean of extracted CIP numeric values per message. \\ \hline
CIP word entropy & Shannon Entropy of parsed CIP fields. For example, a message with 10x4 bytes would have 10 "words". \\ \hline
& \\
\multicolumn{2}{l}{\textit{Exogenous features}} \\ \hline
Day of week & Weekday indicator \\ \hline
Hour of day & Hour indicator \\ \hline
Hour of week & Hour indicator \\ \hline
\end{tabular}
\end{table*}

Because the NetFlow representation transitions from a single scalar feature to a multi-channel feature vector, we additionally included exogenous temporal features. These include hour of day, hour of week, and day of week, which are commonly used in time-series modeling to capture diurnal and weekly periodicities. Such features can improve model confidence and stability, see, e.g, \cite{mohammadi2025deep}.

\paragraph{NetFlow + Payload Dataset}
The 2015 dataset does not provide raw PCAP files for extensive payload extraction; instead, it includes NetFlow records augmented with CIP protocol attributes, more precisely, the encapsulated CIP messages \cite{Maganti2025Discovering}. For the 2017 and 2019 Dec datasets, we deliberately retained the same base feature set, even though richer payload feature engineering would have been possible. This choice ensures comparability of model performance across all datasets.
In the NetFlow+Payload setting, we used the same flow-level feature channels as in the NetFlow-only case and augmented them with payload-derived statistics. These include payload entropies from message and word-level, and payload mean of CIP extracted data.

\paragraph{Training, Calibration, and Sampling}
As the proposed evaluation method relies on conformal prediction, the dataset was split into training, calibration, and test sets using a temporal split of 80/10/10. Feature normalization parameters were computed exclusively in the training set and subsequently applied to the calibration and test sets. Data shuffling was not used because it could allow information leakage from future observations. Similarly, subsampling and folding techniques were avoided. We did, however, sample the data for 10 second aggregates, as we have observed many related works having done earlier. 
During model training, we experimented with different learning rates, embedding dimensions, and time window sizes. We observed no improvement in training or evaluation loss when using embedding dimensions greater than 128 or window sizes greater than 6. Therefore, we opted to keep the complexity of the model to a minimum. The rest of the tunable hyper-parameters are shown in Table~\ref{tab:hyperparams}.

\begin{table*}[t]
\centering
\caption{Hyperparameter search space and functional roles for the proposed graph-temporal neural network model. Some common hyperparameters, e.g., learning rate, are omitted.}
\begin{tabular}{l|c|p{7cm}}
\hline
\textbf{Common Hyperparameters} & \textbf{Value(s)} & \textbf{Description} \\
\hline
Embedding dimension & 128 & Dimensionality of latent node and temporal representations, controlling overall model capacity and attention head size. \\
Graph attention heads & 4 & Number of parallel subspaces used in the multi-head graph attention mechanism. \\
Top-$k$ neighbors & 6 & Maximum number of neighboring nodes attended to per node, controlling graph sparsity and computational cost. \\
Weight decay & $10^{-4}$ & L2 regularization strength applied to model parameters during optimization. \\
\midrule
\textbf{Learnable Hyperparameters} && \\
\midrule
Static prior scale & 10 & Weight of the static graph similarity prior relative to the dynamic context-based similarity. With this parameter, the importance of prior graph can be controlled by initializing it. \\
Attention temperature & 0.9 & Scaling factor controlling the sharpness of the graph attention distribution. \\

\hline
\end{tabular}
\label{tab:hyperparams}
\end{table*}
%
%
%
%
%
%
%
%
%
%
%
%
%
%
%
%
%
%
%
%
%
%
%
\section{Results}
\label{sec:Results}

In this section, we evaluate the performance of our model in comparison with alternative machine learning approaches. We also analyse strategies for selecting the optimal detection threshold and, through illustrative examples, demonstrate how detected anomalies and graph representations reveal the underlying causal relationships. The complete table of results and analysis is provided in the Appendix.

\subsection{Best-Performing Model}
We begin by analyzing the performance of the model in all three data modalities. The proposed STA-GNN method is compared to 
several simpler models in terms of F1-score, FPR, and the number of detected attacks, thereby justifying the model complexity and architectural choices. The results are summarised in Table~\ref{tab:results1}. As an initial model selection strategy, we applied the maximization of the F1-score to determine  decision thresholds for the trained models, which is a common practice in ADS machine learning. The models for comparison include two classical machine-learning methods (K-means and Support Vector Machine (SVM)). Autoregressive models for comparison include Autoencoder (AE) \cite{Goodfellow-et-al-2016}, a unidirectional LSTM-autoencoder (LSTM-AE) \cite{DBLP:journals/corr/MalhotraRAVAS16} and LSTM Variational autoencoder (LSTM-VAE) \cite{DBLP:journals/corr/KingmaW13}. The classical methods were not evaluated for the NetFlow modalities due to their poor performance already in the scalar physical-level model. For the proposed STA-GNN approach, we evaluated two configurations: a simplified variant using only a gated recurrent unit (GRU) without embeddings and temporal attention (STA-GNN*), and the full model incorporating both temporal and spatial attention mechanisms (STA-GNN).

\begin{table*}[ht]
\centering
\small
\setlength{\tabcolsep}{4pt}
\caption{Model comparison across physical and network modalities in terms of
F1-score, false positive rate (FPR), and the number of attacks detected (AD). The classical models with high AD suffer from high FPR, which makes them impractical for realistic deployment scenarios.
The STA-GNN* refers to a simplified variant of STA-GNN without temporal encoding or the temporal attention component.
The best-performing models according to F1-score and AD are highlighted in bold.}
\label{tab:results1}
\begin{tabular}{l|ccc|ccc|ccc}
\toprule
\multirow{2}{*}{Model}
    & \multicolumn{3}{c|}{Physical-level}
    & \multicolumn{3}{c|}{NetFlow}
    & \multicolumn{3}{c}{NetFlow+Payload} \\
& F1 & FPR & AD & F1 & FPR & AD & F1 & FPR & AD \\
\midrule
 K-means
   & 0.29 & 0.829 & 26
   & -- & -- & --
   & -- & -- & -- \\
 SVM
   & 0.24 & 0.860 & 33
   & -- & -- & --
   & -- & -- & -- \\
 Autoencoder
 & 0.72 & 0.003 & 5
 & 0.19 & 0.90 & 36 
 & 0.64 & 0.053 & 8 \\
 LSTM-AE
 & 0.71 & 0.016 & 6
 & 0.20 & 0.88 & 36 
 & 0.73 & 0.003 & 10 \\
 LSTM-VAE
   & 0.72 & 0.001 & 5
   & 0.23 & 0.83 & 35
   & 0.72 & 0.003 & 6 \\
 STA-GNN*
   & 0.74 & 0.002 & 11
   & 0.19 & 0.88 & 35
   & 0.74 & 0.003 & 11 \\
 STA-GNN
   & \textbf{0.77} & 0.004 & 15
   & 0.19 & 0.89 & 36
   & 0.74 & 0.006 & \textbf{16} \\
\bottomrule
\end{tabular}
\end{table*}

Physical-level models with only one scalar feature per node provided the highest F1-score for our models. The two classical machine learning approaches, K-means and SVM, did not produce meaningful results, which is previously observed in in\cite{deng2021graph}. The auto-regressive comparison models, despite their relatively simple structure, achieved an F1-score close to that of the best-performing models. However, a closer inspection of detected attacks shows that its performance is misleading as the models successfully detect only a few attacks. The inflated F1-score is explained by the fact that there is an attack that accounts for more than 40\% of the attack data points. Any model capable of detecting this attack significantly improves 
its F1 score. This observation highlights that strict reliance on F1-score maximization is inadequate to evaluate anomaly detection models in this context.

NetFlow-models without CIP-payload data were not able to reliably detect attacks in any of the studied cases, as they produced excessive false positives, rendering them impractical for deployment. This behavior is likely due to the noisy and low-semantic structure of flow-level data due to NetFlow summarizing traffic using only coarse statistical aggregates. In contrast, incorporating payload information substantially improved performance, as evidenced by the NetFlow+Payload model achieving detection capabilities comparable to the physical-level model. Although the physical-level model produced the lowest false positive rates overall, the NetFlow+Payload configuration detected the largest number of attacks.

\subsection{Nonconformity Scoring and Thresholding}
Table~\ref{tab:results1} demonstrates that threshold selection strategy plays a critical role in practical model performance. Although maximizing the F1-score reduces the FPR, further improvements are possible. By applying difference-based nonconformity scoring, we significantly reduce false positives while at the same time, quite surprisingly, increase the number of detected attacks. The FPR can be treated as a user-defined parameter and set to a desired level through the calibration scores. For the six-day baseline, it was not feasible to enforce guaranties below an FPR of 0.001, as the calibration set is too small and the assumption of exchangeability degrades at more extreme thresholds. Breaking the exchangeability, in turn, leads to poor attack detection performance. Longer and more stable baseline periods would enable stronger guaranties and better align with operational requirements. For example, \cite{ahmed2020challenges} note that even a single false positive every six months can be considered excessive in industrial deployments. This rate corresponds to an FPR on the order of $10^{-6}$—approximately three orders of magnitude lower than the achievable thresholds in our setting.

\begin{table*}[ht]
\centering
\caption{Evaluation results of the STA-GNN model under two thresholding strategies:
F1-maximization and difference nonconformity scoring. Choosing the latter gives highest AD, but with a very low F1-score. We emphasise that F1 does not always reflect the desired performance of the model.}
\label{tab:results_grouped}
\renewcommand{\arraystretch}{1.15}
\begin{tabular}{l l | ccc | ccc}
\toprule
& & \multicolumn{3}{c|}{\textbf{F1-max threshold}}
  & \multicolumn{3}{c}{\textbf{Conformal calibration threshold}} \\
\cmidrule(lr){3-5} \cmidrule(lr){6-8}
\textbf{Dataset} & \textbf{Modality}
& F1$_{\max}$ & FPR$_{\max}$ & AD & F1$_{\text{conf}}$ & FPR$_{\text{conf}}$ & AD \\
\midrule

\multirow{3}{*}{SWaT 2015}
  & Physical-level     & 0.77 & 0.004 & 15 & 0.03 & 0.001 & \textbf{20} \\
  & Netflow      & 0.22  & 0.881 & 36 & 0.01  & 0.001 & 9  \\
  & Netflow+Payload   & 0.74 & 0.006 & 16 & 0.02 & 0.001 & \textbf{22} \\ 
\bottomrule
\end{tabular}
\end{table*}

\begin{figure*}[htpb]
        \subfloat[Attack to AIT202.]{
            \includegraphics[width=.48\textwidth]{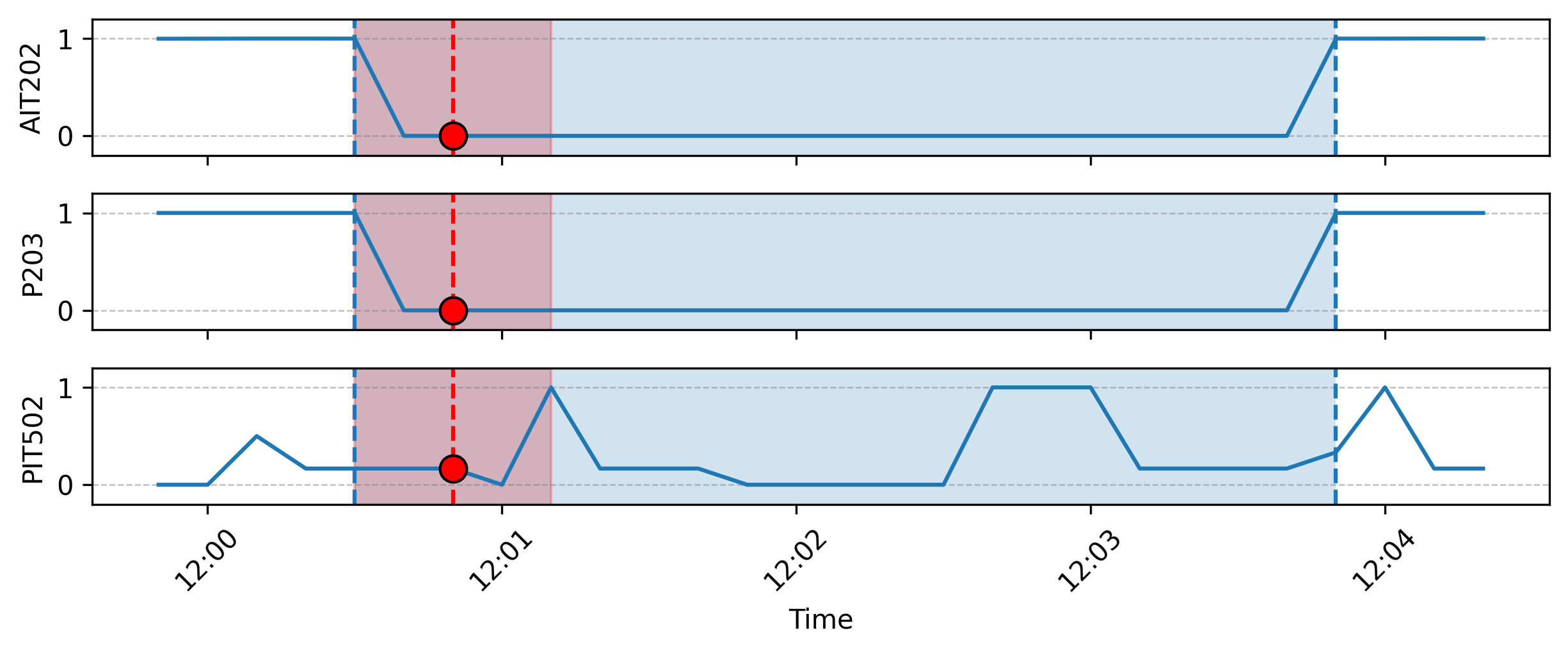}
            \label{fig:testi}
        }
        \hfill
        \subfloat[Attack to DPIT301.]{
            \includegraphics[width=.48\textwidth]{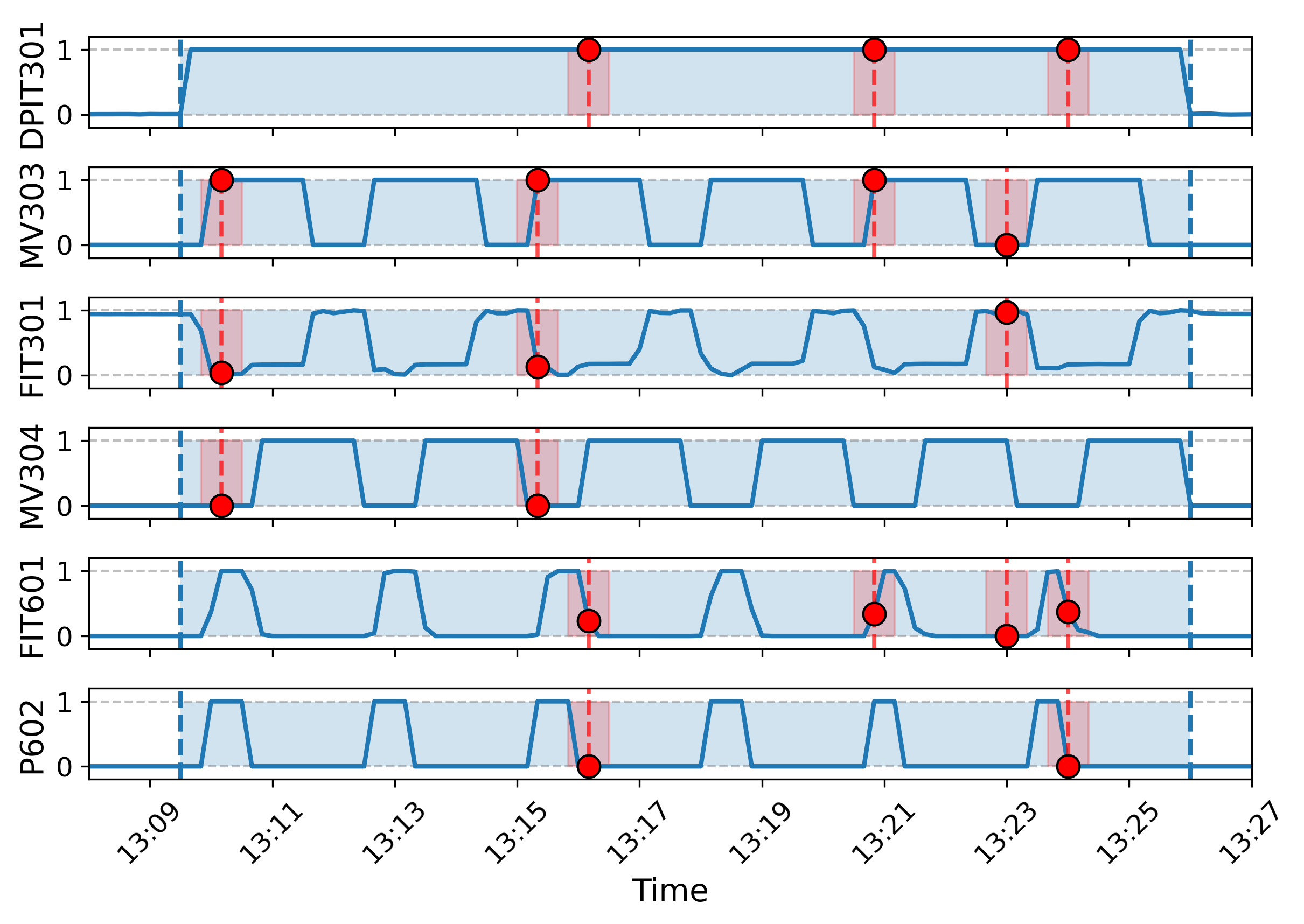}
            \label{fig:testi2}
        }
        \caption{Comparison of normalised sensor response windows (shaded red) during the attack window (shaded blue and separated with blue dashed line). (a) The attack to AIT202 
        was detected only once in the beginning of the attack. (b) The attack to DPIT301 was detected multiple times during the attack, from various sensors and actuators (a cascade failure). For clarity, we only show top 3 anomalous sensors per detected anomaly.}
        \label{fig:side_by_side}
\end{figure*}

Difference-based conformal thresholding also allows the model to adapt to different phases of an attack. Once an alarm is raised, subsequent observations within the same attack episode do not trigger repeated alerts. Although absolute nonconformity scores may remain above the threshold, their relative changes do not, 
effectively suppressing redundant alarms. This behavior provides additional qualitative insight that short, transient attacks tend to trigger a single alert, not affecting much to the system, whereas prolonged or system-wide cascade failures continue to generate alarms, reflecting their severity and the urgency of response. This is demonstrated in Fig.~\ref{fig:testi}, where no cascade-failure occurs. Here the attack has no effect on the system and remains a point source. On the other hand, in Fig.~\ref{fig:testi2}, an attack on a sensor triggers alarms throughout the system during the attack, suggesting cascade failure. We also note that the true source of the attack is not often detected in cascade failures. This is illustrated in 
Fig.~\ref{fig:testi2}, the attack against DPIT301 is detected seven minutes after the attack started because other device reconstruction errors dominate and trigger the alarm elsewhere.

Finally, while difference-based nonconformity scoring reduces false alarms through strict FPR guaranties, it also leads to low F1-scores when evaluated under conformal thresholds. Indeed, the resulting F1-scores fall below 0.04 in all cases, as shown in Table~\ref{tab:results_grouped}. 
Nevertheless, the model remains highly effective at detecting attacks when the decision threshold is set with a conformal evaluation strategy. 

\subsection{Model Performance Across Datasets}
\begin{figure*}[ht]
    \centering
    \subfloat[]{
        \includegraphics[width=.8\textwidth]{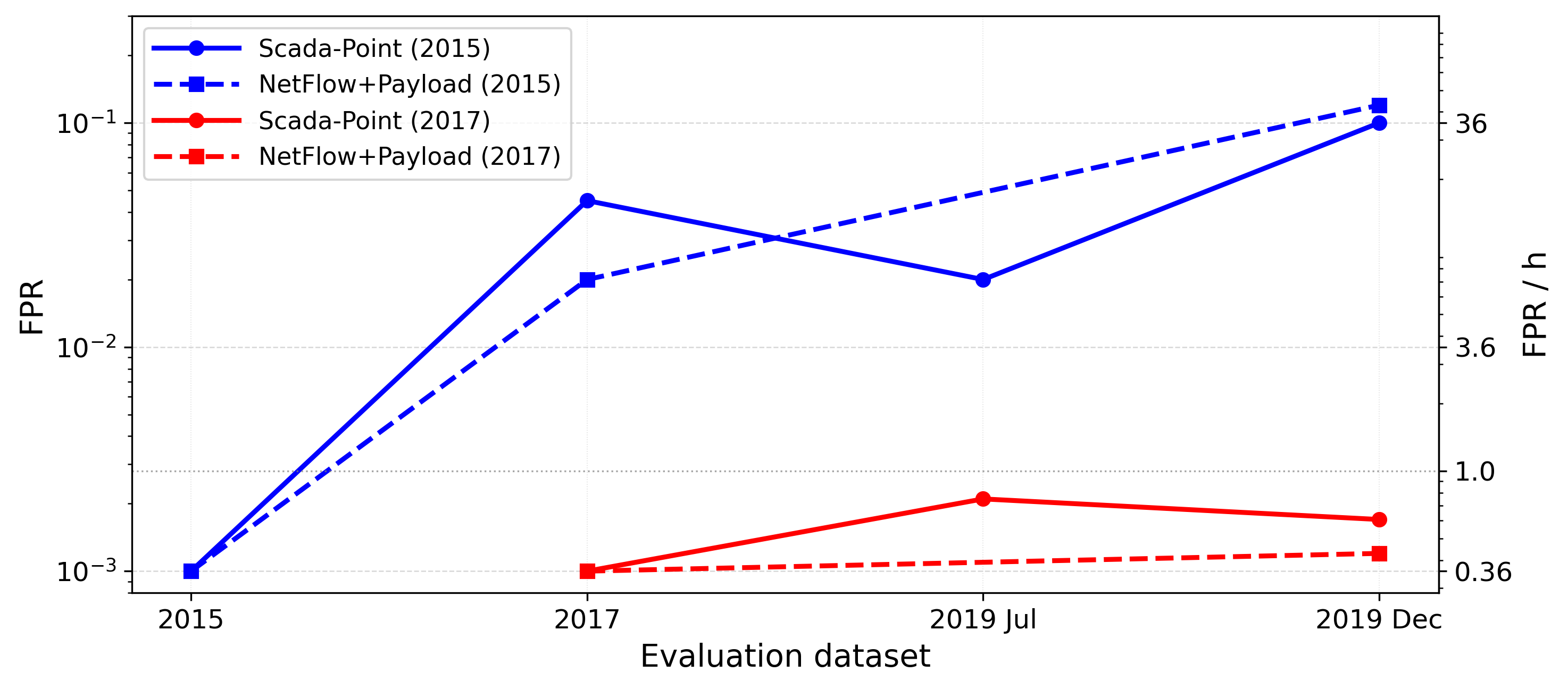}
        \label{fig:top}
    }
    \vspace{0.5cm}
    \subfloat[]{
        \includegraphics[width=.8\textwidth]{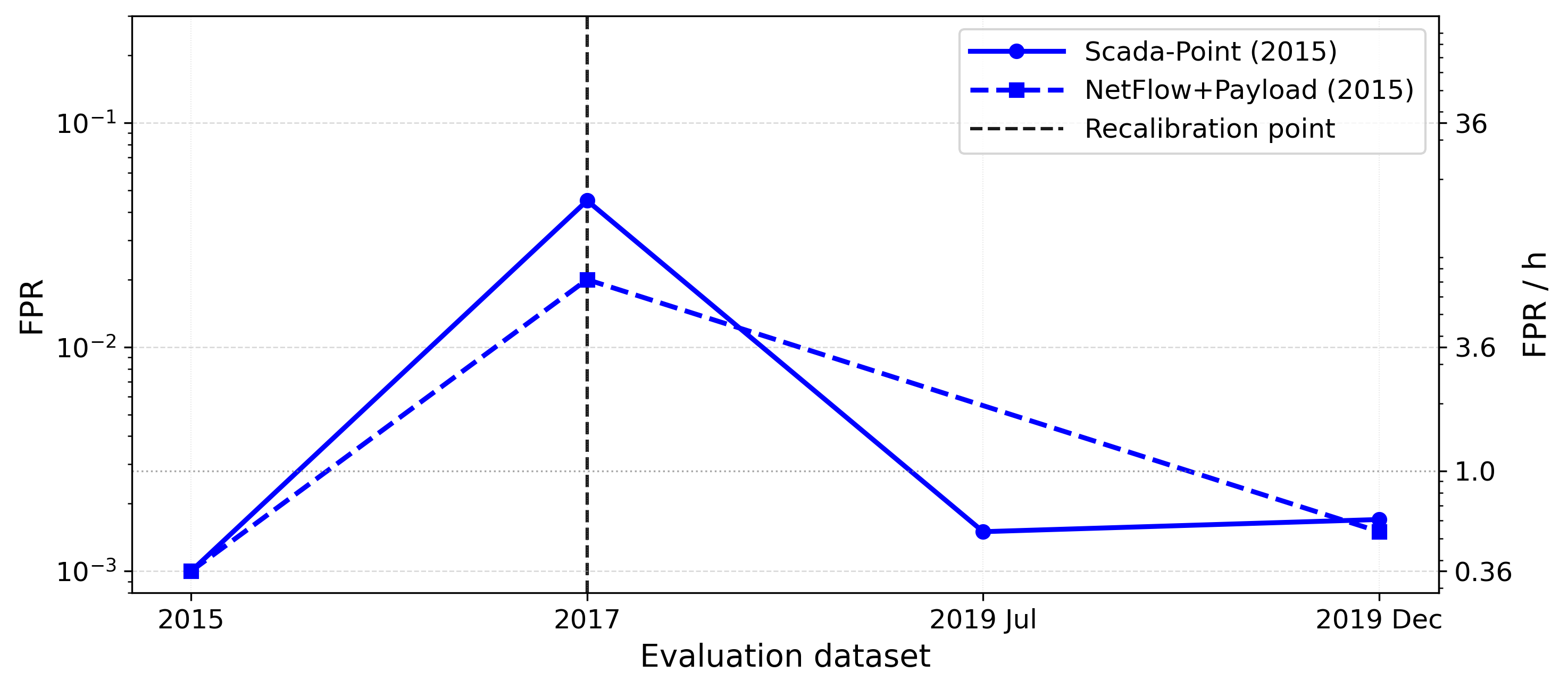}
        \label{fig:bottom}
    }
    \caption{The FPR across datasets. Top (a): Model performance with (red) and without (blue) retraining. Bottom (b): Performance with recalibration of the 2015 model using the 2017 baseline. The FPR can be controlled with recalibration, which is often more feasible than retraining the model.}
    \label{fig:twostack}
\end{figure*}

\paragraph{Physical-level.}
The conformal framework enables explicit control over the FPR, providing monitoring of the model performance over time. Gradual increases in the FPR can serve as indicators of degraded model performance or baseline drift, and this phenomenon is clearly observed in our experiments. As shown in Fig.~\ref{fig:twostack}, the model trained on the 2015 dataset exhibits a sharp performance decline when applied to data from later years. Already in the 2017 dataset, the FPR increases on the order of $10^{-2}$, corresponding to approximately 3--4 alarms per hour, which would be impractical for real-world deployment. This behavior suggests a baseline drift, which aligns findings, for example, in \cite{YANG2026115380}. The results indicate that the model is highly sensitive to even minor shifts in individual sensor signals. Our model repeatedly alerts from sensor AIT201 and few other sensors. Although we did not investigate the signals in detail, we can confidently say that there is a drift as the same sources repeatedly alert.

To further investigate this temporal degradation, we retrained a separate model using the 2017 dataset as a baseline and compared it to the 2019 July and December datasets. In this setting, the physical-point model again fails. However, it now holds the FPR guarantee but is not able to detect attacks effectively. We do not detect similar shift of the sensors that we detected earlier with 2015 model.

Yet another advantage of non-conformal scoring scheme, a topic not covered so far, is the possibility to deal with the baseline drift via recalibrating the scores. The drift occurs because of various reasons, e.g., wearing of the equipment, variations in environmental conditions, sensor aging or recalibration of the equipment. The drift has been observed in SWaT datasets and reported, for example, in \cite{turrin2020statistical}. Recalibration of conformal scores can adjust the decision threshold and prolong the performance of the model, without requiring extensive retraining of the model. Thus, we recalibrated the 2015 model with 2017 data. This time, the model retains it's FPR for 2019 datasets, but unfortunately, could not retain it's anomaly detection capability in this case either. The inefficiency of recalibration is indirect evidence of another type of drifting, i.e., concept drift. Unlike covariate drift, concept drift is a result of change in the testbed configuration. 
In formal terms, the problem space changes. 
In covariate drift, the input space changes, 
which can be dealt with recalibration of the model. For example, changes of data processing pipelines, alterations in operating or usage patterns cause concept drift. In \cite{turrin2020statistical}, the authors further speculate that this could be the case between the 2015 and 2019 SWaT datasets. 

\paragraph{NetFlow+Payload.}
We followed the same evaluation and adaptation strategy for Netflow+Payload modality. Although recalibration and retraining hold the low FPR guarantees, the models detect only 2 out of 6 attacks. This outcome is 
expected, as even in the original 2015 dataset only approximately half of the attacks were detected. The Dec 2019 dataset is thus a poor indicator of model performance because it contains only a few attacks. Furthermore, while recalibration proves sufficient to maintain the FPR guarantees, the squared prediction errors per node and per time step increase in 2019 Dec dataset. This ultimately leads to 
rendering the model impractical for long-term deployment. Consequently, as the observed growth of prediction errors is rather evidence of concept than covariate drift, the retraining of the model remains the most reliable option for model deployment.

Another likely explanation of poor detection rate of attacks is the incompleteness of the original NetFlow data, which we intentionally replicated during the preprocessing of the 2017 and 2019 PCAP files. The primary limitation in this setting arises from the preprocessing and feature representation of the network traffic. More expressive feature engineering, such as incorporating write tags or richer descriptors of payload-level behavior, is likely to
improve detection performance, as demonstrated in \cite{Maganti2025Discovering}. However, a detailed investigation of optimal modeling and feature design in this context lies outside the scope of this work, which focuses on model endurance rather than benchmarking, and is therefore left for future research.

\subsection{The Attention Graphs and Explainability}
\begin{table*}[ht]
\centering
\renewcommand{\arraystretch}{1.2}
\caption{Summary of correct detection and causal inference performance for the SWaT 2015 dataset across Physical-level and NetFlow+Payload modalities. The pure NetFlow modality is excluded because it did not yield meaningful results; see the detailed analysis 
in the Appendix A.}
\begin{tabular}{ccc|ccc}
\hline
\multicolumn{3}{c|}{Physical-level}
& \multicolumn{3}{c}{Netflow+Payload} \\
\cline{1-6}
Alarms Raised & \multicolumn{2}{c|}{Correct}
& Alarms Raised & \multicolumn{2}{c}{Correct} \\
\cline{2-3}\cline{5-6}
 & Detection & Causality
 &  & Detection & Causality \\
\hline
20 & 15 & 12 & 22 & 15 & 14 \\
\hline
\end{tabular}
\label{tab:causalcorrelate}
\end{table*}

The final attention-weight graph $G_a$, together with the highest anomaly scores, enables the inspection of both the anomaly points and their correlations within the system. We examine the detected attacks and their associated attention weights and study how these correlations respond to causal relationships. We use the documented system architecture, known causality maps provided in \cite{maiti2023iccps}, and examples from \cite{wang2022faithful} for qualitative analysis. In Table~\ref{tab:causalcorrelate}, we summarise the findings, with the analysis and rationale provided in the Appendix A. Among the alarms raised in the Physical-level and NetFlow+Payload modalities, approximately 68 to 75\% of attacks were correctly detected and traced, while correct or partially correct causal relationships were identified in approximately 60 to 63\% of the alerts raised.

To improve the clarity of the experimental evaluation, 
we summarise for each attack the target component, detection latency, identified nodes, and whether the detection and inferred causal relationships are correct. For instance, in the DPIT301 attack, the model raises an alarm approximately fourteen minutes after the attack onset, correctly identifying DPIT301 as the affected sensor. In the attention graph, DPIT301 appears as a central hub with the highest cumulative attention, consistent with its role as the compromised sensor. The bidirectional interaction between FIT601 and DPIT301 reflects realistic pressure–flow dynamics, while strong attention from MV301, MV303, and MV304 toward DPIT301 indicates system-level responses to the disturbance. Additional attention assigned to PIT502 from FIT601, DPIT301, and P602 suggests downstream propagation effects rather than root-cause attribution. This example illustrates how attack-level analysis provides insights beyond aggregate scores, distinguishing immediate detection from cascade-driven responses. A compact summary is provided in Table \ref{tab:attack_level_summary}, with full per-attack details in the Appendix \ref{app:Appendix}.  

\onecolumn
\begin{table}[ht]
\centering
\caption{Summary of attack-level detection performance with selected examples. Detection latency is measured from attack start to first alarm. Detection indicates whether at least one identified node corresponds to the true attack source or directly affected component. Causality evaluates whether the attention graph aligns with known system dependencies. A complete per-attack breakdown, including reasoning and full feature attribution, is provided in the Appendix \ref{app:Appendix}.}
\label{tab:attack_level_summary}

\rowcolors{2}{gray!10}{white}
\begin{tabular}{lllllll}
\hline
\textbf{Attack ID} & \textbf{Target} & \textbf{Modality} & \textbf{Latency} & \textbf{Detected Nodes} & \textbf{Detection} & \textbf{Causality} \\
\hline
6  & AIT202 & Physical & Immediate ($\sim$0 min)   & AIT202, P203, PIT502 & Yes & Yes \\
8 & DPIT301 & Physical & Delayed ($\sim$14 min) & FIT601, DPIT301, P602 & Yes & Yes \\
19 & AIT504 & Physical & Short delay ($\sim$5 min) & AIT504, P502, AIT503 & Yes & Partially \\
29 & PLC 0.20 & NetFlow+Payload & Immediate ($\sim$0 min) & -   & No & No \\
30 & \makecell[l]{LIT101, P101,\\ MV201} & Physical & Immediate ($\sim$0 min) & \makecell[l]{FIT201, AIT502, P205,\\ LIT101, AIT202} & Yes & Partially \\
32 & PLC 0.30 & NetFlow+Payload & Immediate ($\sim$0 min) & PLC 0.30 & Yes & Yes \\
41 & LIT301  & Physical & Delayed ($\sim$38 min) & \makecell[l]{LIT301, AIT501, FIT502,\\ FIT601, MV303} & Yes & Partially \\
\hline
\end{tabular}
\end{table}

\begin{figure*}[htbp]
    \centering
    \includegraphics[width=0.4\linewidth]{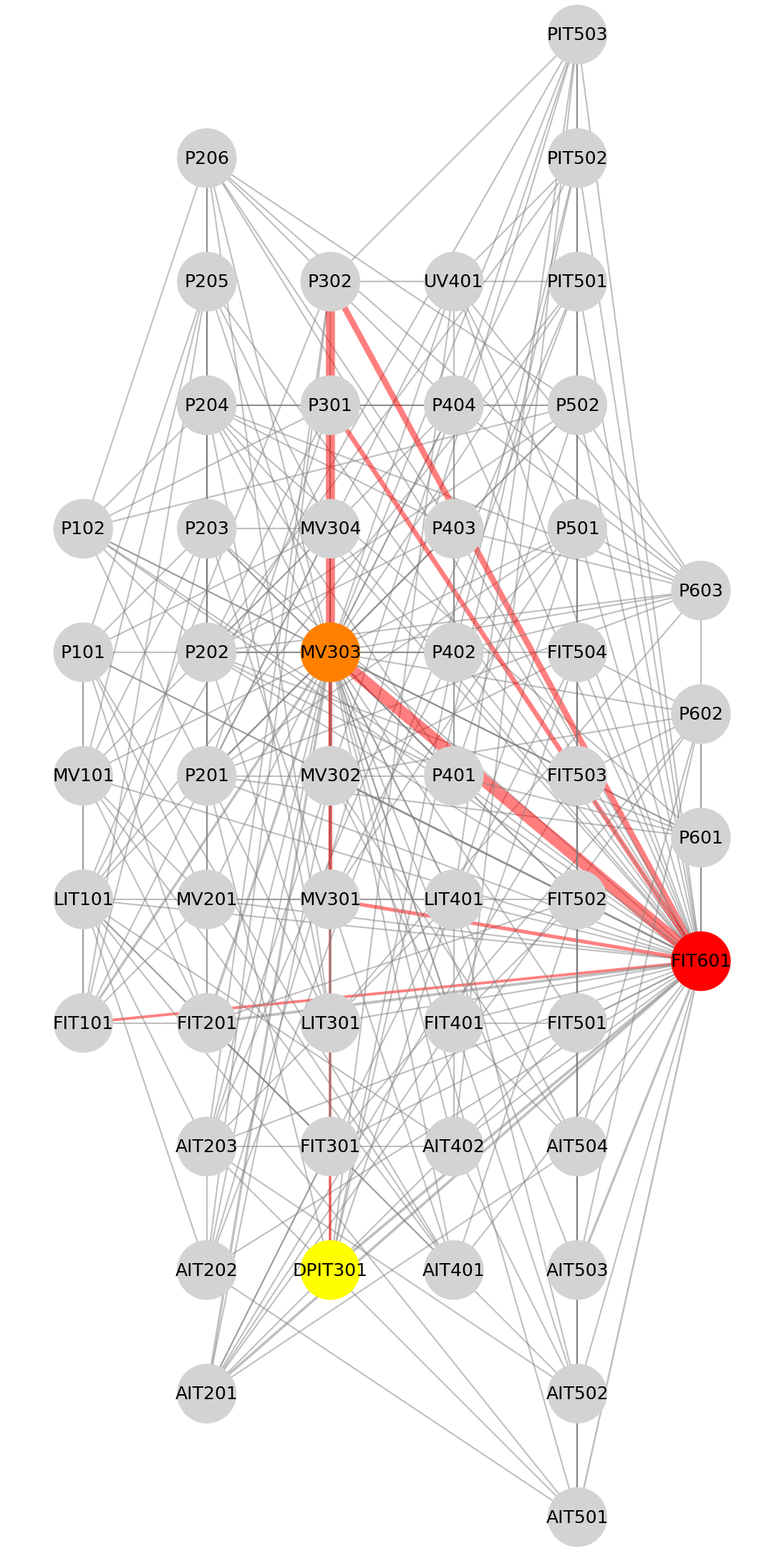}
    \caption{Attack on DPIT301 detected via anomalies in FIT601, with attention edges highlighting system-level dependencies between distant process stages.}
    \label{fig:placeholder}
\end{figure*}

When an alarm is raised, the outcome can be interpreted in two ways: 
whether the detection localises the true source(s) or the immediately affected devices, and
whether the edges of attention reflect the correct underlying causality. These interpretations 
allow us to distinguish between correct detections and meaningful causal explanations. Causality might be captured despite mislocalization, and vice versa. Furthermore, cases in which either the detection or the inferred causal structure or both are incorrect. This analysis raises an important aspect in model evaluation: Attention graphs allow us also to assess whether the model is functioning meaningfully. In highly imbalanced evaluation datasets, containing many attacks within a short time period, the model can simply raise alarms and occasionally “guess” correct results without having converged to a well-functioning representation. This could lead to a false sense of security that the model is functioning correctly. For example, in the NetFlow modality, although nine attacks were detected, we observed that the model recognised them by chance. Alarms were consistently raised on incorrect PLC devices and did not produce meaningful attention edges. As an example of successful model performance, we use a known result that an attack on the backwash (DPIT301) causes malfunctioning of the pumps P601 and P602 \cite{wang2022faithful}. This attack is detected by our model as an anomaly in the flow meter (FIT601) and is illustrated in Fig.~\ref{fig:placeholder} (the same attack as the one illustrated in the sensor level in Fig.~\ref{fig:testi2}). The attention edges with highest weights indeed capture the relationship between these stages, even though they are far apart in the system.

For Netflow+Payload data, i.e., using feature channels and IP addresses as nodes yields the best results when combined with CIP payload data. However, this configuration reduces interpretability and explainability. In fact, we can only trace alarms and attention edges to the IP-level, which is less informative than physical-level representations. We cannot directly identify which physical devices are attacked and we can only trace events back to the PLC-level. Furthermore, the attention edges are not often informative, as many of these relationships are already well-known \emph{a priori}. However, many real ICS environments are highly complex and may include hundreds of PLC devices and workstations. As the system size increases, this method becomes increasingly feasible and valuable.

\begin{figure*}[htbp]
    \centering
    \subfloat[No Prior Graph]{
        \includegraphics[width=.4\textwidth]{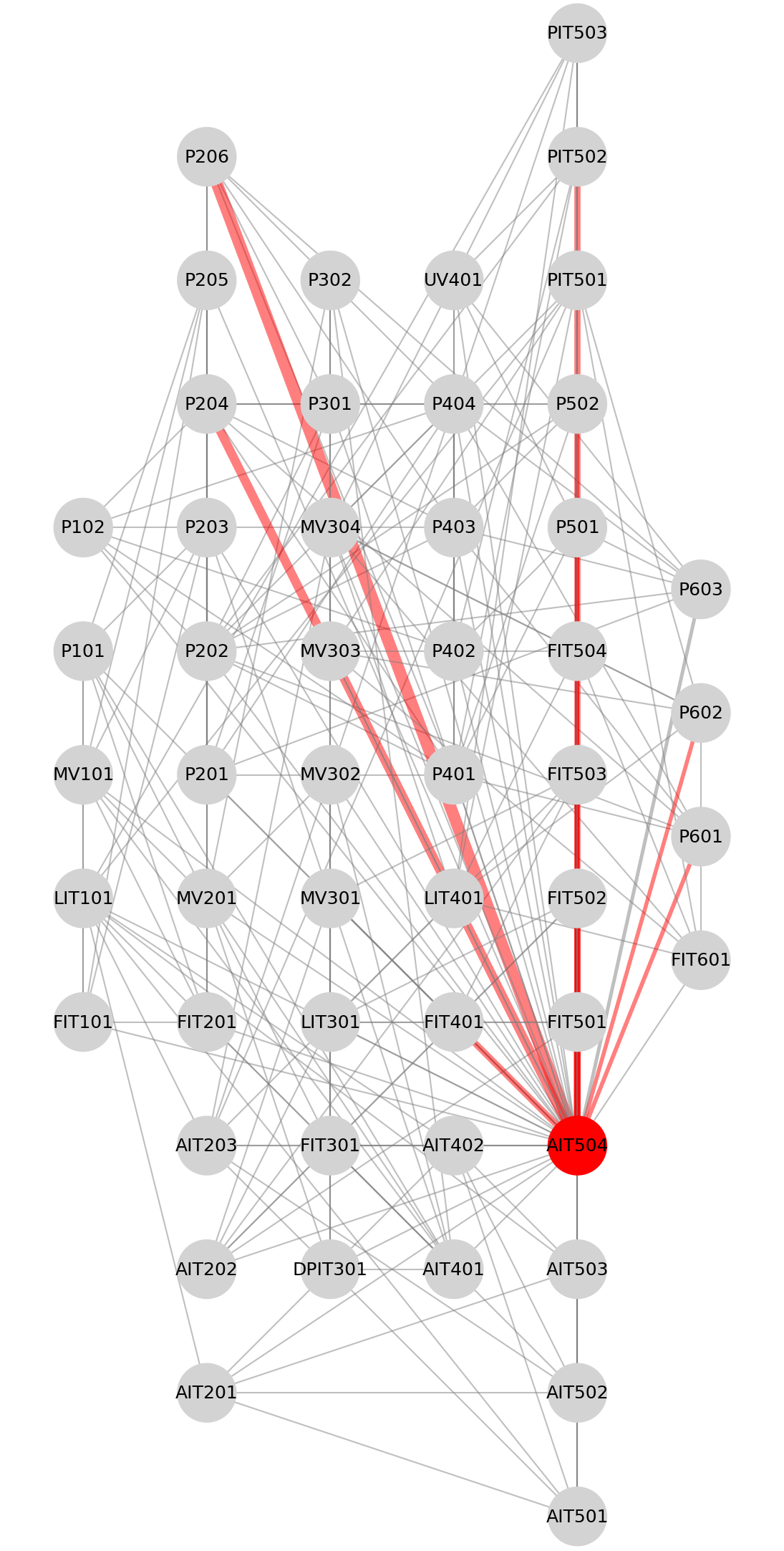}
        \label{fig:first}
    }
    \hfill
    \subfloat[Prior Graph]{
        \includegraphics[width=.4\textwidth]{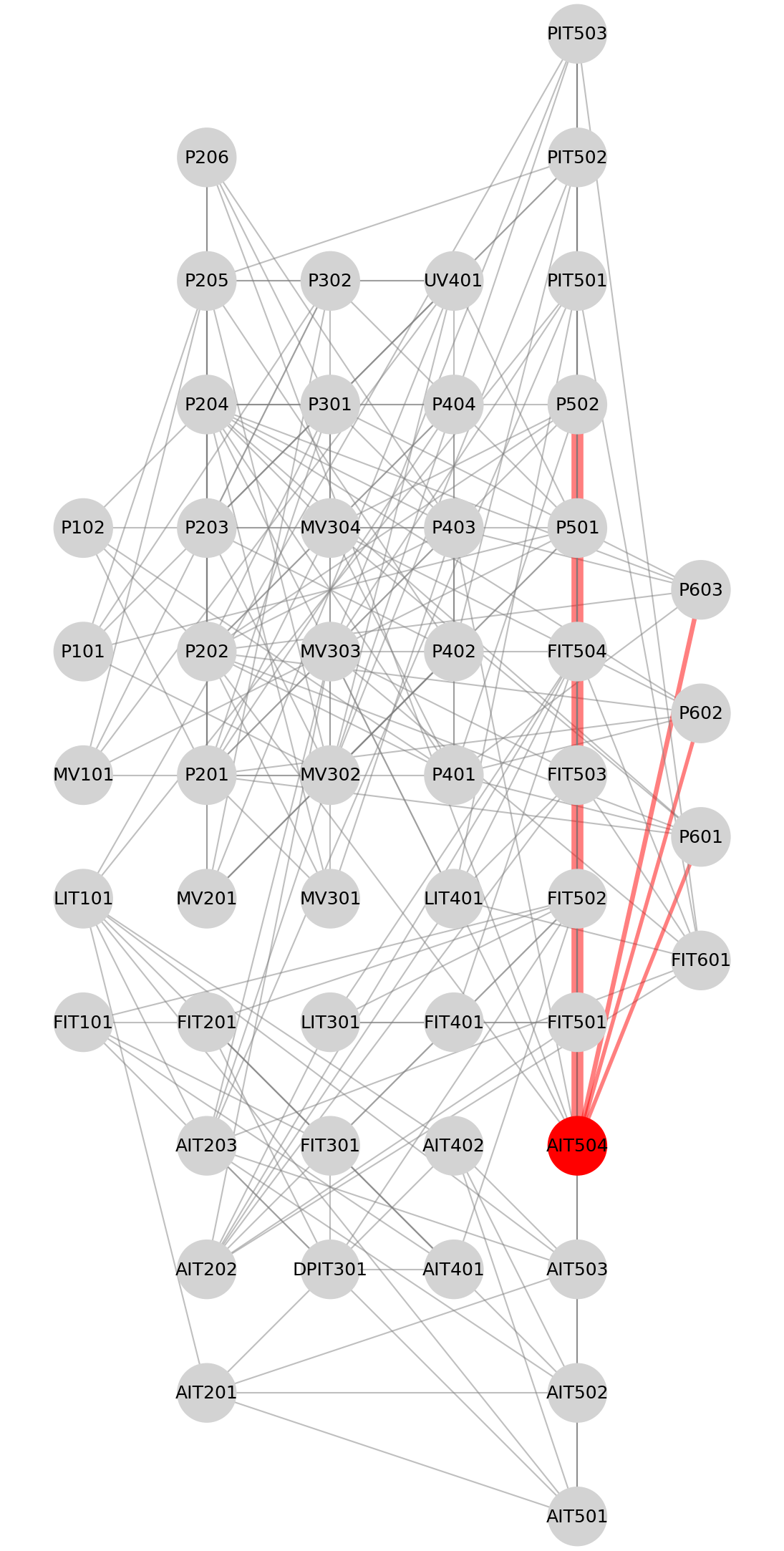}
        \label{fig:second}
    }
    \caption{Attack to AIT504 without (a) and with (b) 
    soft prior graph. The soft prior helps filtering the edges that are not related to causality. The grey edges in the background are the contextual learned edges + static graph from temporal attention. Only spatial attention edges from anomalous nodes are retained for clarity. Note that the detected anomaly points are also removed, 
    because the prior restricts the dynamical similarity.}
    \label{fig:both}
\end{figure*}

Finally, we show by comparison how prior knowledge of the system shapes the attention edges. The prior structure is derived from the adjacency graph presentation of the system, such that components within the same stage are considered fully connected. The connecting components are then linked to other processing stages, as deduced from the system description in \cite{SWATORIG}. This is only one example, and alternative prior graph constructions such as causal directed graphs have been investigated, for example, in \cite{maiti2023iccps}. In Fig.~\ref{fig:first}, without structural constraints, the inferred causal relationships in the model can be dominated by noisy correlations. For example, pumps or valves that exhibit similar behavior are often connected by attention edges, even if they are physically far apart in the system and no true causal connection exists. When an alarm is raised, edges connected to correlated but non-causal devices may reduce the practical usefulness of the model. The resulting graph with a stronger prior is in Fig.~\ref{fig:second}. The meaningless correlations are no longer present.

We retained a simple prior graph for two reasons: (a) we are non-experts in the system domain and lack detailed operational expertise, and (b) we wanted to allow the model to learn the structure autonomously, rather than letting the prior dominate. This approach enables the detection of long-range dependencies in an \textit{ad hoc} manner, as illustrated in Fig.~\ref{fig:placeholder}. Finally, we note that using strong prior knowledge of the system does not necessarily improve detection accuracy, as it may reduce long-range dependencies; however, it can enhance the explainability. This trade-off will be explored in future work.
%
%
%
%
%
%
%
%
%
%
%
%
%
%
%
%
%
%
%
%
%
%
%
%
\section{Discussion}
\label{sec:Discussions}

In this section, we examine the methodological and practical issues encountered during our analysis and reflect on how our findings agree or deviate 
from previous work in the literature. Here, we focus on the limitations of commonly used evaluation schemes, the operational relevance of our results, and the broader challenges of applying machine learning in industrial cybersecurity. We will also critically assess our modeling choices, including the role of explainability, architectural constraints, and multimodal inputs. 
These reflections will shed light on the limitations of our approach and discuss the directions in which future work should focus on to achieve reliable and deployable anomaly detection in real-world systems.

A central challenge in evaluating anomaly detection models for cyber-physical systems is that commonly reported metrics, most notably the F1 score, do not always reflect the true operational value of the model. One reason is that the duration of an attack heavily influences the F1 score, but many anomaly detection models detect an attack only after it has begun to significantly 
affect the system. However, the early stages of an attack often cause negligible physical deviation, which makes them difficult to detect. Penalising the model for not recognizing these weak initial signals results in a lower F1 score even when the model performs exactly as required in practice, i.e. alerting when the system deviates from normal behavior. This discrepancy leads to misleading comparisons in the literature, where the number of detected attacks is rarely reported. Our results in Table~\ref{tab:results_grouped} underscore the problem that the F1 score might be very low but performs better than using the F1 maximizing strategy. The other aforementioned benefit from nonconformity scoring support using it as thresholding method and as a framework.

Event-based F1 evaluation has been proposed, where each detected attack is flagged as a single positive instance. This would make the model comparison more uniform. However, this does not necessarily make the F1 score more representative of the performance of the model, as the imbalance due to the attack durations still biases the metric. A persistent issue is that long system-wide attacks can dominate the score. In the SWaT 2015 dataset, for example, an attack ID~28 (see Appendix \ref{app:Appendix}) is relatively easy to detect because it targets the pump P302 and triggers a cascade failure across the system. Correct detection of this single attack accounts for approximately 60\% of all anomalous time points. As a result, any model that identifies this event achieves a substantially inflated F1 score. We observed that auto-regressive models detected only a few attacks, yet their F1-scores were almost comparable to that of our best-performing model. Moreover, in most model-design studies we reviewed, the number of detected attacks is often not explicitly reported. This limits the interpretability of benchmark comparisons, as we argue that, along with FPR, the ability to detect a diverse set of distinct attacks is a critical factor in assessing practical model performance.

High FPR is another key 
issue in anomaly detection. Frequent false alarms tend to 
impose a high operational burden, leading to fatigue of alerts and reducing operator trust in the system. A custom is that a useful model is trained with the lowest possible FPR, even at the expense of a lower detection rate of true anomalies. This is also a limitation of our model such that we rather keep the FPR low and allow some attacks to remain undetected. Furthermore, manual inspection shows that a substantial portion of false positives are directly followed by attacks and related to them. Removing these attack-adjacent alerts from false positives reduced our FPR count by 40\% in the physical-level model, leaving only a small number of genuinely spurious alarms. This is yet another indication that operational relevance is not always captured by standard metrics.

The issues discussed so far reflect a broader challenge in machine-learning-based cybersecurity research, in which many published models are evaluated primarily under benchmark-oriented settings. The emphasis on marginal improvements in recall and accuracy is often a consequence of ambiguous or inconsistent evaluation methodologies. As shown in \cite{arp2021dosdonts}, data leakage, inappropriate sampling, model selection bias from cross-validation, and temporal snooping are widespread pitfalls, particularly in time-series scenarios. Neglecting these issues can lead to overly optimistic performance estimates. Several methods we reviewed report near-perfect F1 scores of 1.0, and some machine-learning approaches claim extremely high detection rates, e.g., those in \cite{kravchik2018detecting,aboah2022anomaly}. We explicitly designed our pipeline to mitigate the risks, for example, by ensuring that no temporal information from the test period is used during training or preprocessing. Although this conservative approach reduces performance on current benchmark datasets, it could yield more reliable estimates for unseen data, which is a critical requirement for deployment in real systems. Therefore, our focus has been on qualitative and causal evaluation of the detected attacks, rather than reporting recall or accuracy.

In addition, we discuss a fundamental issue that is often neglected in many model approaches, namely the 
covariate and concept drifts. The gradual change in the statistical properties of data and changed configurations from time to time 
cause the anomaly detection models to lose accuracy as the system behavior evolves \cite{problem}. We could tackle the covariate shifts with a recalibration approach, but the concept drift always requires re-training of the model. This is an issue for most static machine learning models, where the problem space is unknown. We acknowledge this and admit that the nonconformity scoring does not solve all the problems in dynamic environments but can extend the lifespan of the model. We note also that for model performance observations over time, the monitoring of FPR is an excellent tool, allowed by nonconformity scoring.

Next, we address the known limitation that the attention-based methods are inherently unreliable due to noisy correlations unrelated to causality (see, for example, \cite{shin2025faithful}). In our physical-level modality, the introduction of structural priors significantly reduced spurious attention. The attention mechanism continued to function between physically meaningful components, while irrelevant edges were largely suppressed. This shows that attention mechanisms can be effective when guided by sensible inductive biases. In turn, the method might filter out meaningful, explainable edges as well, which can be considered as a limitation. The prior use is thus a trade-off between interpretability and explainability.

Our analysis of the NetFlow+Payload modality further suggests that incorporating prior knowledge of the system is likely necessary. A small and highly interconnected system representation makes causal interpretation difficult. Because most components appear densely connected at the network level, it becomes challenging to distinguish true process dependencies from generic communication patterns. As a result, although anomalies detected typically rise from correct devices, the attention edges are much more difficult to interpret. This reduced explainability can therefore limit the reliability of causal validation in small environments. In contrast, when the system is larger and contains more distinct components, the richer structural variability typically makes causal patterns easier to isolate. This allows dependencies, propagation paths, and abnormal interactions to become more clearly distinguishable than in a small $\sim$10-component network like SWaT testbed. Confirming this hypothesis in larger and more realistic industrial control system environments remains as an important direction for our future research.

Finally, some recent work argues that effective detection of industrial anomalies requires combining payload information with netflow data \cite{alsoscada,Pinto2023Survey}. We did find evidence supporting this claim. For 2015 dataset, we could find 26 attacks when combined the two methods (20/22 separately). We remind, however, that the Netflow model requires the Payload data for the model to work properly, which increases the model complexity and computation needs. However, it should be noted
that the physical-point detection model is typically simple and easily importable after SCADA-point. The netflow+payload detection might be difficult for encrypted data, as 
the data before SCADA point is often secured and owned by system vendors \cite{cis2024ics}, which limits 
the practical deployability of such approaches in operational environments.
%
%
%
%
%
%
%
%
%
%
%
%
%
%
%
%
\section{Concluding Remarks}
\label{sec:Conclusions_FutureWork}

In this study, we have proposed a Spatio-Temporal Attention Graph Neural Network (STA-GNN) for multi-purpose anomaly detection in industrial control systems. The model produces explainable graph-based attention graphs that enable the investigation of system behavior. By incorporating prior knowledge of the system, these attention mechanisms can be used to detect anomalies and reason about their potential consequences.

Beyond model design, this work highlights several fundamental challenges in applying machine learning to industrial control systems, to which our approach is also subject. A key issue is the gap between model development and real-world deployment. In practice, the objective is not to train a theoretically optimal model but rather to deploy a system that reliably detects attacks while minimizing false alarms. Our results demonstrate that commonly used evaluation strategies, such as maximization of the F1-score, may not capture this operational objective.

We further show that covariate and concept drifts are significant challenges in ICS anomaly detection. Even widely used benchmarking datasets exhibit non-stationarities that render stationary models ineffective over time. To address this, we advocate frequent model recalibration, retraining, and continuous monitoring of performance degradation through false positive rate tracking, enabled by conformal prediction framework. This approach not only ensures operational feasibility, but also provides early indicators of model drift.

Our experiments indicate that the proposed model performs best when applied to physical-point data, while also remaining applicable to NetFlow+Payload-based representations. Although network-level features reduce explainability, they offer improved efficiency. Based on these findings, we recommend a multimodal deployment strategy, combining both physical-level and NetFlow+Payload data to balance interpretability and scalability.

As future work, we aim to integrate the learned attention structures with large language models (LLMs) to further enhance explainability, particularly for non-expert users. By combining attention-based graph representations with facility context and model outputs, such systems could automatically generate human-interpretable explanations and annotations. Ultimately, this direction may enable more intelligent and self-interpreting human–machine interfaces in industrial environments.
\newpage
\section*{Appendix: Analysis of the Attention Weights}
\label{app:Appendix}

\begin{figure*}[ht]
\centering
\includegraphics[width=\textwidth]{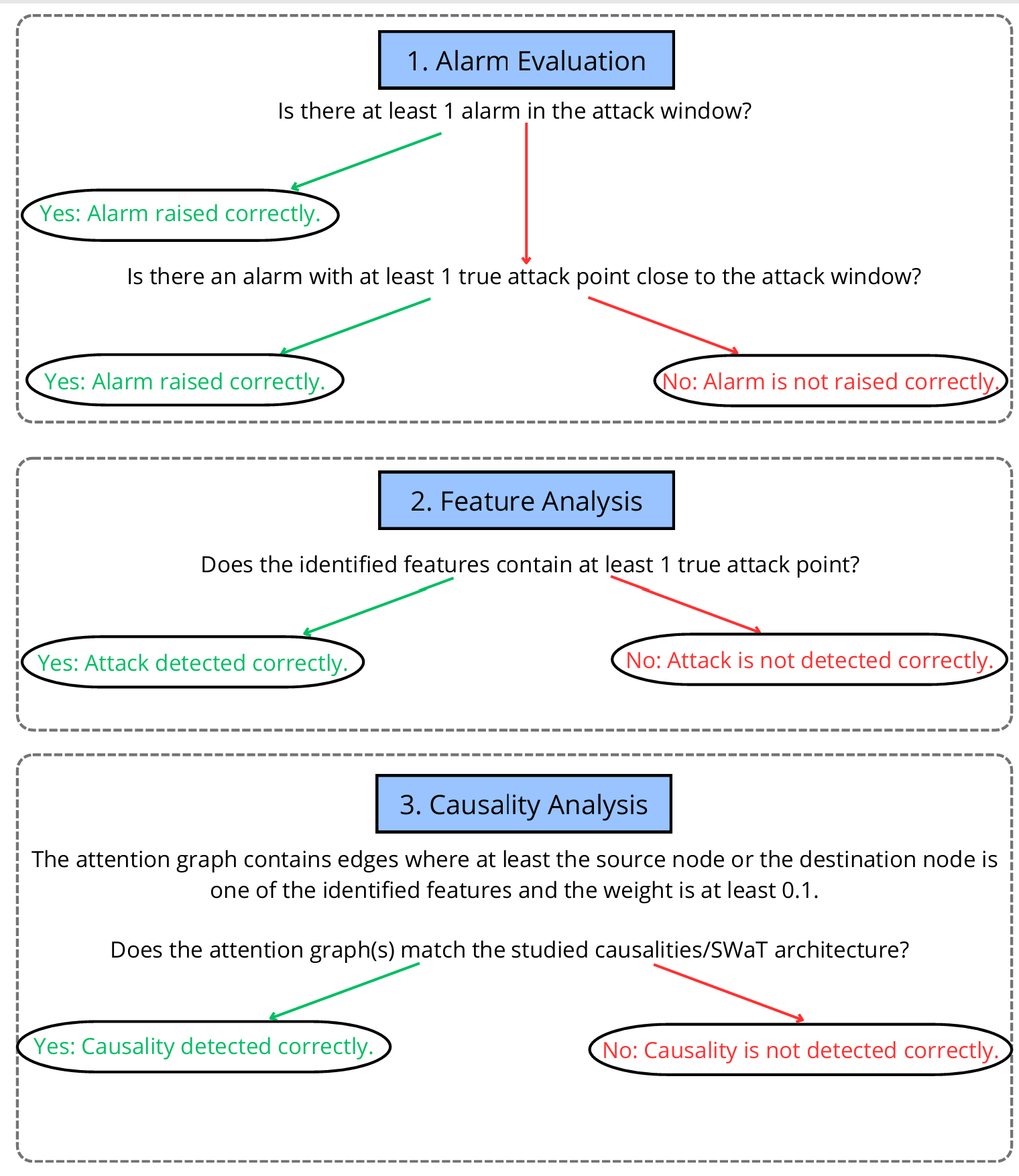}
\caption{Analysis Pipeline}
\label{fig:Kirsin}
\end{figure*}

The analysis of the results of the 2015 model using SWaT 2015 physical dataset consists of three sequential evaluation stages designed to assess alarm quality, feature relevance, and causal validity of attack detection. The graph describing 
the analysis pipeline is illustrated in Fig.~\ref{fig:Kirsin}. The first stage verifies whether an alarm is correctly triggered within (or close to) the attack window. If at least one alarm occurs during the attack window, the alarm is considered to be correctly raised. If not, we check whether there is at least one alarm close to the attack window that corresponds to at least one true attack point. If this condition is met, the alarm is still considered correct. Otherwise, the alarm is classified as incorrectly raised.
The second stage evaluates whether the identified features truly correspond to the attack. The model gives the top 3 features per alarm that have the largest contributions. If the selected features include at least one true attack point, the attack is considered correctly detected. If none of the identified features correspond to true attack points, the detection is considered incorrect (false positive).

The final stage analyses whether the detected relationships are causally meaningful. Attention graphs are constructed using edges 
for which either the source or destination node is among the identified features and the edge-normalised weight is at least 0.1. These graphs are then compared against known 
causal relationships of the system. If the learned attention graph aligns with the expected causal structure, causality is considered correctly detected. This means that the edge directions match known causal relations, the involved nodes correspond to components known to influence each other, and the relation is documented in the literature or consistent with SWaT architecture. If the attention graph has nodes unrelated in architecture, cross-stage connections with no physical/control dependency, edges contradict a known process, or random high-weight edges flow, the causality is considered incorrectly detected. 
The causality can also be considered partially correct in one of the following situations: correct nodes but wrong direction, indirect but valid path, subsystem-level match, weak but meaningful edge, or partial feature overlap. The first 
is a situation in which a correct dependency is identified, but the directionality is incorrect. This suggests that the model captures the dependency but not the causal direction. When the path is valid but indirect, the model captures higher level dependency but skips the intermediate node. This may indicate abstraction or shortcut learning. In subsystem-level matches, the model identifies correct process region but not the exact documented pairs. If an edge matches known causality but is much weaker than unrelated edges, the signal exists, but the model does not strongly prioritise it. This indicates that the edges are meaningful, but they are too weak. If there exists partial future overlap, only one node in the edge is part of the true attack chain, but the other is only strongly related in the architecture. This means that the model captures the attack region but not the exact causal pair. Also, some inferred edges appear plausible given system dynamics, but cannot be conclusively validated against documented process architecture or literature. These relations are therefore categorised as partially detected causality rather than confirmed physical causal chains.

Table~\ref{tab:appendix} contains the results of the analysis of the alarms raised by the 2015 model using the SWaT 2015 physical dataset. The table does not include attack numbers 5, 9, 12, 15, and 18 because they do not cause physical impact on the system. The table contains the attack time, attack description, detected features with largest contribution, 
alarm quality assessment, feature relevance, and causal validity, as well as details about the results for each attack. The column with the attack time contains the date and the true attack window. The attack description has the true information about the attack as well as the expected impact or attacker intent. The columns Alarm Raised, Detected Correctly, and Causality Detected Correctly contain the evaluation results explained above. The detected features summarises all the top 3 features identified by the model inside or near the true attack window. The Details column tries to explain the reasoning of the evaluation results. It describes the attention graphs, states the raised alarms, and identified true attack points.
Table~\ref{tab:KostinTaulu} contains similar analysis results using the Netflow+Payload modality. In the Netflow+Payload modality, the much smaller and highly interconnected system representation makes causal interpretation difficult. Because most components appear densely connected at the network level, it becomes challenging to distinguish true process dependencies from generic communication patterns. As a result, although anomalies detected typically rise from correct devices, the attention edges are much more difficult to interpret. This reduced structural transparency can therefore limit the reliability of causal validation in small environments, even when anomaly detection performance itself remains reasonable. In contrast, when the system is larger and contains more distinct components, the richer structural variability typically makes causal patterns easier to isolate, allowing dependencies, propagation paths, and abnormal interactions to become more clearly distinguishable than in a small ~10-component network. It remains future work for us in a larger, realistic ICS environment.

\onecolumn
\begin{landscape}
{\footnotesize
\begin{longtable}[c]{||c|c|m{1.4cm}|m{2.4cm}|m{.8cm}|m{1.2cm}|m{1.2cm}|m{1.4cm}|m{10cm}||}
\caption{SWaT2015 Physical.\label{tab:appendix}}\\
\hline\hline
Model & Attack & Attack Time & Attack Description & Alarm Raised & Detected Correctly & Causality Detected Correctly & Detected Features & Details \\
\hline\hline
2015  & 1 & 28/12/2015 10:29:14 - 10:44:53  & Open MV101. Tank overflow. & No & - & - &  & No alarms raised before the second attack.   \\ \hline
2015  & 2 & 28/12/2015 10:51:08 - 10:58:30 & Turn on P102. Pipe bursts. & Yes & No & Yes & FIT601, MV303, MV301 & The alarm raised at 10:52:30 falls within the ground-truth attack interval and therefore meets the temporal criterion for correct detection. However, the set of alarmed features does not directly include the true attack point, P102. Inspection of the attention graph shows that FIT601 exhibits strong outgoing attention toward P101, P102, MV301, MV302, and PIT502. In addition, MV303, MV301, and MV302 form strongly connected patterns, with evident internal subsystem coupling between MV303 and MV302, as well as between MV301 and MV201. Although both P101 and P102 receive attention, P102 is explicitly present as a destination node. This suggests that the attack on P102 primarily manifested through altered pump behavior, leading to observable flow deviations captured by FIT601, which in turn triggered control responses reflected in MV301 and MV303.  \\ \hline
2015  & 3 & 28/12/2015 11:22:00 - 11:28:22 & Increase by 1 mm every second. Tank Underflow; Damage P101. & No & - & - & - & No alarms raised between 10:52:30 and 12:00:40. \\ \hline
2015  & 4 & 28/12/2015 11:47:30 - 11:54:08 & Open MV504. Halt RO shut down sequence; Reduce life of RO. & No & - & - & - & The alarm at 12:00:40 occurs after the attack window and does not involve the true attack point MV504. Given the temporal mismatch and lack of architectural relevance among the alarmed features, the alarm is considered a false positive and the attack a missed detection.  \\ \hline
2015  & 6 & 28/12/2015 12:00:55 -12:04:10 & Set value of AIT202 as 6. P203 turns off; Change in water quality. & Yes & Yes & Yes & AIT202, P203, PIT502 & The alarm raised at 12:00:40 occurs immediately prior to the labeled attack interval (12:00:55–12:04:10) and correctly includes the true attack point AIT202. Analysis of the attention graph indicates that anomalies originating in water quality measurements (AIT202) propagate to pump behavior at P203. Multiple upstream components (P201–P206, FIT201, MV201, and AIT201) exhibit strong attention toward P203, identifying it as a downstream aggregation point in the process flow. Furthermore, PIT502 receives substantial attention from a wide range of upstream flow, pump, and analyser variables (e.g. MV101, P302, and AIT401), consistent with pressure variations arising as downstream effects rather than serving as the root cause of the anomaly.  \\ \hline
2015  & 7 & 28/12/2015 12:08:25 - 12:15:33 & Water level increased above HH. Stop of inflow; Tank underflow; Damage P301. & No & - & - & - & No alarms raised between 12:00:40 and 13:12:40.   \\ \hline
2015  & 8 & 28/12/2015 13:10:10 - 13:26:13 & Set value of DPIT as >40kpa. Backwash process is started again and again; Normal operation stops; Decrease in water level of tank 401. Increase in water level of tank 301. & Yes & Yes & Yes & FIT601, DPIT301, P602 & The alarm raised at 13:24:00 occurs well within the ground-truth attack interval and explicitly includes DPIT301, the true attack point. In the attention graph, DPIT301 emerges as a central hub, receiving the strongest cumulative attention, which is consistent with its role as the injected attack target (a differential pressure sensor). The bidirectional coupling between FIT601 and DPIT301 reflects realistic pressure–flow interactions within the hydraulic process. Additionally, strong attention from control valves MV301, MV303, and MV304 toward DPIT301 indicates system responses to abnormal pressure conditions. Finally, PIT502 receives attention from FIT601, DPIT301, and P602, consistent with downstream pressure propagation rather than serving as the initiating cause of the anomaly.  \\ \hline
2015  & 10 & 28/12/2015 14:16:20 - 14:19:00 & Set value of FIT401 as <0.7. UV shutdown; P501 turns off. & Yes & Yes & Yes & FIT401, PIT502, PIT501, FIT504, FIT503 & The first alarm at 14:16:10 is raised approximately 10 seconds before the onset of the labeled attack, while the second alarm at 14:18:50 occurs well within the attack window. In both cases, the true attack point FIT401 is explicitly included among the alarmed features. Analysis of the first attention graph indicates that FIT401 acts as an upstream driver, with anomalies propagating to P402 and UV401 as local process effects, and to PIT501 and PIT502 as downstream pressure responses. The observed coupling between PIT501 and PIT502 is consistent with pressure equalization rather than a causal initiation. In the second attention graph, FIT401 remains a dominant influence, with strong interactions observed among FIT401, FIT503, and FIT504. As these sensors measure physically adjacent flow streams, this pattern is consistent with hydraulic redistribution. Additional influences from AIT502, LIT101, and PIT503 on FIT504 appear as secondary effects. \\ \hline
2015  & 11 & 28/12/2015 14:19:00 - 14:28:20 & Set value of FIT401 as 0. UV shutdown; P501 turns off. & Yes & Yes & Partially & FIT401, FIT504, PIT501, FIT503, AIT502, & The model raises five alarms (14:19:30, 14:19:40, 14:27:40, 14:28:00, and 14:28:50), and the true attack point, FIT401, is explicitly included in all of them. The attention graphs show that FIT401 exhibits strong outgoing attention toward nearby and related process variables, including FIT504, FIT501, FIT502, LIT401, and AIT402. This pattern is consistent with FIT401 acting as an upstream driver, with anomalies propagating to neighboring flow measurements (FIT50x and FIT504), eliciting a tank level response (LIT401), and affecting associated subsystem measurements (e.g., AIT402). Strong coupling between FIT504 and FIT503, together with links involving PIT501 and PIT503, further supports a physically plausible hydraulic redistribution and downstream pressure response: once flow is perturbed at FIT401, adjacent flow and pressure signals reorganise and become mutually predictive. At later stages of the attack, AIT502 emerges as a prominent hub, with many variables attending to it and strong coupling observed between AIT502 and AIT501. This behavior may reflect a secondary downstream water-quality effect developing over time, although it could also indicate a broader attention focus arising from cross-window aggregation or increased sensitivity of the analyser channel. Consequently, causal relationships involving AIT502 are plausible as downstream or secondary effects, but they are less directly attributable to FIT401 than the observed flow and pressure couplings.\\ \hline
2015 & 13 & 29/12/2015 11:11:25 - 11:15:17 & Close MV304. Halt of stage 3 because change in the backwash process. & No & - & - & - & No alarms between 11:08:30 and 12:20:30.   \\ \hline
2015 & 14 & 29/12/2015 11:35:40 - 11:42:50 & Do not let MV303 open. Halt of stage 3 because change in the backwash process. & No & - & - & - & Alarm was raised 38 minutes after the attack. Despite the overlapping feature MV303, the delayed-effect explanation is unlikely. Repeated alarms occurring long after the attack window indicate persistent false positives or sensitivity to normal process dynamics. \\ \hline
2015 & 16 & 29/12/2015 11:57:25 - 12:02:00 & Decrease water level by 1mm each second. Tank Overflow. & No & - & - & - & The alarm at 12:20:30 occurs well after the attack window (11:57:25–12:02:00) and does not include the true attack point LIT301. Despite involving related P3 variables (MV303, FIT301, DPIT301), the temporal mismatch indicates the alarm is a false positive.   \\ \hline
2015 & 17 & 29/12/2015 14:38:12 - 14:50:08 & Do not let MV303 open. Halt of stage 3 because change in the backwash process. & Yes & Yes & Yes & MV303, FIT601, P602 & The alarm raised at 14:49:50 occurs well within the ground-truth attack interval and explicitly includes the true attack point, MV303. The attention graph reveals strong coupling from MV303 to MV301, indicating control-level interactions between valves that are consistent with coordinated control logic or shared operational objectives. In addition, strong attention from FIT601 to MV101, MV301, P402, AIT504, and FIT502 suggests that flow deviations propagate downstream to influence multiple actuators and sensors. Attention from P602 to MV101, MV301, and P402 further indicates pump-driven propagation affecting valve states and pressure-related components. Finally, attention from DPIT301 to MV303 reflects pressure-based feedback influencing valve behavior, which is characteristic of closed-loop control. Overall, this pattern aligns with the expected propagation of a valve-actuator attack in the SWaT system: manipulation of the valve directly alters flow, subsequent flow and pressure changes trigger pump and valve responses, and pressure sensors (e.g., DPIT301) provide feedback to the control logic.\\ \hline
2015 & 19 & 29/12/2015 18:10:43 - 18:15:01 & Set value of AIT504 to 16 uS/cm. RO shut down sequence starts after 30 minutes. Water should go to drain. & Yes & Yes & Partially & AIT504, P502, AIT503 & The alarm raised at 18:15:20 explicitly includes the true attack point, AIT504. In the attention graph, links from AIT504 to P501, P502, and PIT502 are consistent with water-quality anomalies being associated with downstream actuation and pressure behavior, or at least co-varying with these variables. Attention from P502 to P501, PIT502, and MV101 reflects a plausible pump-driven propagation pattern in which pump activity influences valve states and pressure responses. Additional attention from AIT503 to P501, P502, and FIT504, as well as from AIT502 to AIT503, suggests coupling among analyser channels and their interaction with hydraulic variables. However, the attention graph is relatively sparse and the inferred influence is not fully centered on AIT504. While the graph supports a process-consistent pattern of propagation and dependency, it is more appropriate to interpret these relationships as influence pathways rather than as evidence of strict physical causation originating from AIT504 and propagating to pumps.\\ \hline
2015 & 20 & 29/12/2015 18:15:43 - 18:22:17 & Set value of AIT504 to 255 uS/cm. RO shut down sequence starts after 30 minutes. Water should go to drain. & No & - & - & - & No alarms raised between 18:15:20 - 18:43:40.  \\ \hline
2015 & 21 & 29/12/2015 18:30:00 & Keep MV101 on countinuosly; Value of LIT101 set as 700 mm. Tank overflow. & No & - & - & - &  No alarms raised between 18:15:20 - 18:43:40.  \\ \hline
2015 & 22 & 29/12/2015 22:55:18 - 23:03:00 & Stop UV401; Value of AIT502 set as 150; Force P501 to remain on. Possible damage to RO. & Yes & Yes & Partially & FIT504, FIT503, PIT501, FIT401, AIT502 & The model raises alarms at 22:55:00, 22:55:10, 22:55:40, 22:55:50, and 23:02:40. While none of the alarms explicitly include the true attack points UV401 or P501, the final alarm correctly identifies AIT502. This behavior is consistent with the physical attack propagation in the system: manipulation of the UV401 valve alters the flow dynamics, which is first reflected in flow sensors FIT503 and FIT504, subsequently impacts tank behavior and pressure measurements (PIT501 and P501), and eventually affects the analyser AIT502. Accordingly, the attention graphs predominantly highlight FIT503, FIT504, PIT501 and FIT401 during the early and mid stages of the attack, and AIT502 in the later stage. The root actuator UV401 is not explicitly highlighted, which is likely due to its binary, low-variance nature and its role as an actuator rather than a continuously varying sensor.\\ \hline
2015 & 23 & 30/12/2015 01:42:34 - 01:54:10 & Value of DPIT301 set to >0.4 bar; Keep MV302 open; Keep P602 closed. System freeze. & Yes & Yes & Partially & DPIT301, P602, MV301, FIT601, MV303 & The model raises two alarms at 01:53:40 and 01:54:20, both of which correctly identify the true attack point DPIT301. The first alarm additionally captures the true attack point P602. In the corresponding attention graph, DPIT301 is strongly emphasised and exhibits clear connectivity with P602. The graph also assigns high attention to MV302 (e.g., through strong DPIT301 to MV302 and MV302 to DPIT301 edges). However, the features selected by the model include MV301 instead of the true attacked actuator MV302. In the second attention graph, the attention shifts toward MV303 and FIT601, while DPIT301 remains central. The strong coupling observed between FIT601 and MV303 suggests the presence of downstream propagation or symptom dynamics toward the end of the attack, rather than the model explicitly isolating the original attacked actuator MV302. \\ \hline
2015 & 24 & 30/12/2015 09:51:08 - 09:56:28 & Turn of P203 and P205. Change in water quality. & No & - & - & - & No alarms raised between 06:59:40 and 11:14:40.   \\ \hline
2015 & 25 & 30/12/2015 10:01:50 - 10:12:01 & Set value of LIT401 as 1000; P402 is kept on. Tank underflow. & No & - & - & - & No alarms raised between 06:59:40 and 11:14:40. \\ \hline
2015 & 26 & 30/12/2015 17:04:56 - 17:29:00 & P101 is turned on continuosly; Set value of LIT301 as 801 mm. Tank 101 underflow; Tank 301 overflow. & Yes & No & Partially & MV303, FIT301, P602 & The model raises an alarm at 17:21:40 but does not explicitly identify either of the true attack points, P101 or LIT301. However, the attention graph reveals a clear edge from LIT301 to MV303, indicating that LIT301 is included within the model’s attention neighborhood. In addition, P602 exhibits strong connectivity to P101, providing an indirect link to one of the true attack points. The features highlighted by the model, namely FIT301 and MV303, represent plausible intermediate variables that are influenced by changes in tank level and pumping behavior. Together, these observations suggest that the model is capturing the downstream propagation effects of the attack rather than directly isolating the root attacked sensors. This behavior is consistent with the physical characteristics of the system, where LIT301 is a level sensor with relatively slow dynamics and P101 is a pump whose impact is more readily observed through flow and valve-related measurements. \\ \hline
2015 & 27 & 31/12/2015 01:17:08 - 01:45:18 & Keep P302 on contineoulsy; Value of  LIT401 set as 600 mm till 1:26:01. Tank overflow. & No & - & - & - & No alarms raised between 00:14:40 and 02:22:40.   \\ \hline
2015 & 28 & 31/12/2015 01:45:19 - 11:15:27  & Close P302. Stop inflow of tank 401. & Yes & No & Yes & PIT501, PIT503, FIT401, FIT503, FIT501, AIT501, AIT402, FIT301 & The model raises 11 alarms at 02:22:40, 02:22:50, 02:23:20, 02:23:30, 02:24:10, 02:24:30, 02:24:50, 02:25:10, 08:10:40, 08:18:20, and 11:15:30. None of the early alarms explicitly include the attack point P302. In the initial attention graphs, the dominant nodes are FIT401, FIT501, FIT503, PIT501, and PIT503, with strong attention from LIT401 to FIT401 and pronounced coupling between flow and pressure variables. This indicates that the attack has begun, but its impact is distributed across the hydraulic subsystem and is partially mitigated by control loops. During the middle stage of the attack, the model raises repeated clusters of alarms involving similar feature sets. The corresponding attention graphs exhibit stable interaction patterns, particularly between FIT401 and FIT501, FIT501 and FIT503, and PIT501 and FIT503. These stable attention structures suggest that the system is operating in an abnormal yet controlled regime, where controllers compensate for the disturbance and thereby obscure the underlying root cause. As a result, attention patterns stabilise rather than drift over time. In the later stage of the attack, new dominant nodes emerge, including AIT402, AIT501, MV303, MV304, and P302. The attention graphs at this stage show a many-to-one convergence toward AIT402, strong bidirectional connections, and pronounced actuator–sensor feedback loops. Water quality variables increasingly integrate the accumulated hydraulic stress, leading the process into a globally inconsistent state. At this point, the influence of the attack origin P302 becomes statistically unavoidable and is explicitly reflected in the attention structure.\\ \hline
2015 & 29 & 31/12/2015 15:32:00 - 15:34:00 & Turn on P201; Turn on P203; Turn on P205. Wastage of chemicals. & No & - & - & - & The alarm at 15:47:40 occurs well after the attack window (15:32:00–15:34:00) and therefore cannot be attributed to the attack. Although one alarmed feature (P205) overlaps with an attack point, the substantial temporal delay indicates the alarm is not raised correctly.  \\ \hline
2015 & 30 & 31/12/2015 15:47:40 - 16:07:10 & Turn P101 on continuously; Turn MV101 on continuously; Set value of LIT101 as 700 mm; P-102 started itself because LIT301 level became low. Tank 101 underflow; Tank 301 overflow. & Yes & Yes & Partially & FIT201, AIT502, P205, LIT101, AIT202 & The model raises alarms at 15:47:40 and 16:07:20. The second alarm correctly highlights LIT101, which is one of the true attack points. In the first attention graph (15:47:40), two dominant hubs emerge: FIT201 and AIT502. Both exhibit multiple outgoing edges with moderate-to-high attention weights and are strongly connected to pumps and analysers (e.g., AIT203, AIT202, P203, and P205). FIT201 functions as a flow integrator positioned between actuators and tanks; it aggregates downstream effects and typically responds earlier than level sensors. At the onset of the attack, LIT101 is manipulated, leading to inconsistencies in MV201/P101 control actions. As a result, flow control mechanisms react before deviations in tank level become fully observable. Consequently, the model initially captures a hydraulic or flow imbalance rather than a direct level anomaly. The prominence of AIT502 indicates concurrent shifts in water quality and analyser dynamics. Changes in flow affect residence time, causing chemical signals to lag behind hydraulic responses while still reacting relatively quickly. The presence of non-trivial attention weights from AIT503, PIT502, and AIT201 to AIT502 further suggests that the anomaly has already propagated across multiple subsystems and is no longer localised. In the second attention graph, strong attention is concentrated on LIT101, with significant incoming contributions from MV303, AIT402, and AIT201. This pattern indicates that the model now treats LIT101 as a central state bottleneck, requiring multiple control and analyser nodes to explain its behavior. MV201 appears only indirectly in this graph, acting as a contributing node rather than a central hub. The relatively lower attention weights associated with MV201 are consistent with its binary nature, its frequent masking by control logic, and its limited continuous expressiveness compared to sensor measurements. \\ \hline
2015 & 31 & 31/12/2015 22:05:30 - 22:11:40 & Set LIT401 to less than L. Tank overflow. & No & - & - & - & No alarms raised between 22:01:50 and 22:31:50.   \\ \hline
2015 & 32 & 1/1/2016 10:36:00 - 10:46:00 & Set LIT301 to above HH. Tank underflow; Damage P302. & Yes & Yes & Yes & LIT301, AIT202, FIT301 & The model raises an alarm at 10:47:20, correctly flagging LIT301, the true attack point. The corresponding attention graph exhibits strong incoming attention to AIT202 from multiple components, including LIT301, MV201, FIT201, P101, P203, and P205. This pattern indicates that the model has identified AIT202 as a bottleneck state that most effectively explains the anomalous system behavior. As an analyser sensitive to residence time and flow redistribution, AIT202 integrates slow but cumulative effects, making it a natural point at which the consequences of level manipulation manifest. In addition, the attention graph highlights strong influence from LIT301 to AIT202, MV201, and P402, reflecting flow redistribution, pressure–flow coupling, and control responses induced by the level inconsistency. Importantly, LIT301 appears upstream in the attention flow, exerting influence over both actuators and analysers. This indicates that LIT301 is not merely reacting to downstream disturbances but actively driving system dynamics, providing a strong signal of causal relevance rather than simple correlation. \\ \hline
2015 & 33 & 1/1/2016 14:21:12 - 14:28:35 & Set LIT101 to above H. Tank underflow; Damage P101. & Yes & Yes & Partially & FIT601, P602, MV301, LIT101, MV101, FIT101 & The model raises two alarms, at 14:23:00 and 14:29:40. The first alarm does not directly include LIT101 and therefore does not, by itself, provide a clean identification of the attack’s root cause. In contrast, the second alarm explicitly includes LIT101, the true attack point, along with MV101 and FIT101, which are tightly coupled to Tank 1 dynamics. The first attention graph is characterised by strong edges between FIT601 and MV303 (bidirectional), as well as notable interactions involving P602, MV301, and MV201. Additional, weaker links connect FIT601 to P602, P603 to P602, and AIT504 to P602. This structure is not centered on LIT101; instead, it forms a compact cluster around FIT601 and MV303 and a set of pump and valve interactions (P602, MV301, MV201). This pattern suggests that the disturbance originating in Tank 1 induces broader control adjustments across the plant, leading the model to highlight a secondary inconsistency in another unit. Such behavior is consistent with SWaT system dynamics, where control actions intended to stabilise one subsystem can introduce transient disturbances in others. At this stage, the model correctly detects abnormal behavior, but the attention is focused on a correlated downstream subsystem rather than the true origin of the attack. The second attention graph, however, shows a markedly different and more localised structure. Strong edges appear from LIT101, MV101, and FIT101 to P101, along with additional influence from MV101 and FIT101 to MV201 (and P203). In this graph, P101 emerges as the central sink, receiving the strongest incoming attention. Its behavior is primarily explained by valve state (MV101), flow (FIT101), and tank level (LIT101), which closely reflects the physical operation of the Tank 1 control loop. Importantly, LIT101 appears upstream in the influence flow, indicating that it is driving the observed inconsistencies rather than merely responding to downstream effects. Overall, this attention pattern ties the anomaly directly to the Tank 1 actuation pathway (LIT101–MV101–FIT101–P101), which is precisely where level spoofing or manipulation is expected to manifest as control inconsistency.
\\ \hline
2015 & 34 & 1/1/2016 17:12:40 - 17:14:20 & Turn P101 off. Stops outflow. & No & - & - & - & No alarms raised between 16:23:00 and 17:19:00.   \\ \hline
2015 & 35 & 1/1/2016 17:18:56 - 17:26:56 & Turn P101 off; Keep P102 off. Stops outflow. & Yes & No & Partially & FIT601, P602, MV301 & The model raises an alarm at 17:19:00, which falls within the attack window, but the set of alarmed features does not include the true attack points P101 or P102. Instead, the attention graph highlights a strong coupling between the flow sensor FIT601 and the valve MV303. This indicates that the anomaly is expressed through their interaction: either MV303 exhibits unusual behavior that is reflected in FIT601, or FIT601 becomes anomalous and MV303 is identified as the actuator most closely associated with flow regulation in that context. From a physical process perspective, the more plausible causal direction is from MV303 to FIT601, as valve actuation directly influences flow measurements. In addition, the attention graph reveals a strong interaction between pressure (P602) and valve MV301, as well as a strong association between MV301 and another actuator, MV201. Both P602 and MV301 are also linked to the analyser/transmitter AIT202, suggesting that deviations in pressure–valve dynamics propagate to other dependent measurements. These relationships imply that, under anomalous conditions, the involved variables become mutually informative and that the model localises the inconsistency within this tightly coupled control loop. Several weaker connections are also observed, including links from P603 and AIT504 to P602, and from P301, P302, MV302, and MV304 to MV301. Taken together, these patterns suggest that although the attack is initiated at P101/P102, its effects propagate downstream and manifest most prominently within a pressure–flow–valve regulation loop, where inconsistencies are detected earlier and more strongly by the model. \\ \hline
2015 & 36 & 1/1/2016 22:16:01 - 22:25:00 & Set LIT101 to less than LL. Tank overflow. & No & - & - & - & The first alarm at 22:15:00 occurs before the attack window (22:16:01–22:25:00) and does not include the true attack point LIT101, while the second alarm at 22:26:00 occurs after the attack has ended and also omits the attack point. Consequently, neither alarm can be attributed to the attack.   \\ \hline
2015 & 37 & 2/1/2016 11:17:02 - 11:24:50 & Close P501; Set value of FIT502 to 1.29 at 11:18:36. Reduced output. & Yes & No & Partially & FIT504, FIT503, FIT401, PIT501 & The model raises four alarms at 11:17:50, 11:18:30, 11:18:40, and 11:19:10. None of these alarms directly identify the true attack points, P501 or FIT502. Instead, across all timestamps, the attention graphs consistently reveal a dense subgraph centered on flow-related sensors. In particular, strong couplings are observed between FIT503 and FIT504, FIT503 and FIT502, FIT503 and FIT501, FIT503 and PIT501/PIT502/PIT503, FIT504 and PIT501/PIT502, as well as between FIT401 and P402/UV401. These patterns indicate that the model identifies FIT503 and FIT504 as the primary carriers of anomalous information, with many other sensors becoming informative because their readings are no longer consistent with the observed flow behavior. In effect, the anomaly manifests as a breakdown in the normal flow–pressure relationships within the subsystem. A plausible propagation pattern is therefore that the attack initially manipulates P501 and FIT502, leading to changes in upstream and downstream flow distribution. The resulting inconsistencies are detected first in FIT503 and FIT504, which are highly sensitive and respond rapidly, while pressure-related anomalies (e.g., PIT501) emerge at a later stage. However, the attention graphs do not provide a clear causal direction from pumps or valves to flow and subsequently to pressure.   \\ \hline
2015 & 38 & 2/1/2016 11:31:38 - 11:36:18 & Set value of AIT402 as 260; Set value of AIT502 to 260. Water goes to drain because of overdosing. & Yes & Yes & Yes & AIT502, AIT402, LIT401, AIT202, & The model raises alarms at 11:32:10, 11:33:00, 11:37:00, and 11:37:30. Importantly, both true attack points, AIT402 and AIT502, are present in every alarm. Across all timestamps, the attention graphs consistently identify AIT502 as a central hub, exhibiting strong outgoing connections to AIT402, AIT202, P501, and FIT503, as well as a strong bidirectional coupling with AIT402. Beyond this core structure, the attention graphs reveal secondary propagation to LIT401, FIT401, P402, UV401, and FIT503, with FIT503 becoming increasingly prominent at later timestamps. This indicates that the anomaly is primarily analyser-driven rather than flow-driven, with the attacked analysers dominating the attention structure both early and persistently. Such behavior aligns closely with the ground-truth attack configuration. Moreover, the temporal sequence of alarmed features reflects a plausible propagation pattern: AIT402 and AIT502 become inconsistent immediately following manipulation, related analysers, such as AIT202, and level measurements, such as LIT401, are affected subsequently, and flow-related variables, notably FIT503, emerge later as downstream effects.  \\ \hline
2015 & 39 & 2/1/2016 11:43:48 - 11:50:28 & Set value of FIT401 as 0.5; Set value of AIT502 as 140 mV. UV will shut down and water will go to RO. & Yes & Yes & Partially & FIT401, AIT502, MV101, AIT501, FIT601 & The model raises alarms at 11:44:30, 11:45:10, and 11:45:20, with both true attack pointsFIT401 and AIT502 appearing prominently in all alarms. In the first attention graph, key connections include edges from FIT401 to P402, UV401, and AIT202, from AIT502 to P501 and FIT503, and from MV101 to AIT202, P101, and MV201. This structure indicates that the model associates FIT401 with its local filtration and UV-related subsystem, while AIT502 is linked to a downstream flow–pressure neighborhood. The involvement of MV101 suggests control or actuation activity within the same abnormal regime, consistent with typical plant responses when sensor relationships deviate from normal behavior. In the subsequent attention graphs, strong inbound influence to FIT401 is observed from AIT201, AIT402, and AIT503, alongside persistent strong connections from AIT502 to P501 and FIT503. Additional links involving FIT601 also emerge, including connections to AIT202, FIT502, MV101, and P101. This evolution suggests that the anomaly transitions from a localised disturbance to a system-wide inconsistency, causing a broader set of sensors and actuators to become informative. The appearance of FIT601 is consistent with downstream measurement contradictions that arise as the attack persists. Overall, the attention graphs effectively capture how the anomaly propagates through the system and which dependencies are disrupted. However, they do not reliably establish the underlying physical causal chain, particularly in the presence of controller actions that alter system behavior in response to the disturbance.
\\ \hline
2015 & 40 & 2/1/2016 11:51:42 - 11:56:38 & Set value of FIT401 as 0. UV will shut down and water will go to RO. & Yes & Yes & Partially & FIT401, AIT503, AIT501, FIT504, FIT503, FIT501, PIT501 & The model raises alarms at 11:51:40, 11:52:20, 11:52:30, 11:53:10, 11:53:20, 11:54:00, and 11:54:10, with the true attack point, FIT401, appearing in every alarm. Across all timestamps, the attention graphs are strongly centered on FIT401, which exhibits prominent connections to AIT402, AIT201, and LIT401, reflecting its association with the local measurement and analyser neighborhood, as well as to P402 and UV401, corresponding to the filtration and UV subsystem. Over time, the attention structure expands outward from FIT401 to include FIT504, FIT503, FIT501, PIT501, and other pressure-related variables. This indicates that the anomaly is initially localised at FIT401 and subsequently propagates through downstream flow and pressure relationships, a pattern that is physically plausible for an attack targeting a flow sensor. Consistent with this interpretation, early alarms primarily emphasise FIT401 and analyser-related variables such as AIT503 and AIT501, while mid-phase alarms introduce FIT504 and FIT503, indicating emerging flow inconsistencies. In the later alarms, the inclusion of PIT501 reflects the onset of pressure-related contradictions as the disturbance persists. Overall, the observed progression aligns well with expected process dynamics under sustained abnormal operation. However, the attention graphs do not reliably establish a strict physical causal chain, particularly in the presence of closed-loop control actions that can alter or obscure apparent cause–effect relationships.  \\ \hline
2015 & 41 & 2/1/2016 13:13:02 - 13:40:56 & Decrease value by 0.5 mm per second. Tank overflow. & Yes & Yes & Partially & LIT301, AIT501, FIT502, FIT601, MV303 & The model raises alarms at 13:41:10 and 13:42:00, consistently identifying LIT301 as a key anomalous node, which corresponds exactly to the true attack point. In the first attention graph, strong coupling is observed between LIT301 and AIT201, along with prominent edges from LIT301 to MV303, FIT502 to P501, AIT501 to P501/P502/PIT501, FIT502 to MV101/P502, and FIT601 to FIT502. This structure suggests that the anomaly is anchored at the level measurement LIT301, with inconsistencies propagating into a surrounding flow–pressure subsystem. In the second attention graph, key edges include connections from FIT601 to AIT202, from LIT301 to MV303 and AIT201, MV303 to AIT202, MV201, MV302, and P302, and from AIT201 to LIT301. This pattern indicates the emergence of a control-loop interaction in which LIT301 influences MV303, which in turn affects other actuators and process variables. Downstream measurements such as FIT601 and AIT202 then become involved as a result of the same underlying regime shift. Overall, the attention graphs plausibly capture the propagation of the anomaly through the system. However, they do not provide a reliable indication of physical causal direction, and the observed structure is more consistent with post-attack recovery dynamics than with the initial onset of the attack itself.
\\ \hline\hline
\end{longtable}
}
\end{landscape}

\onecolumn
\begin{landscape}
{\footnotesize
\begin{longtable}[c]{||c|c|m{1.4cm}|m{4.6cm}|m{1cm}|m{1.4cm}|m{1.4cm}|m{8cm}||}
\caption{SWaT2015 Netflow+Payload.\label{tab:KostinTaulu}}\\
\hline\hline
Model & Attack & Attack Time & Attack Description & Alarm Raised & Detected Correctly & Causality Detected Correctly & Details \\
\hline\hline
2015 & 2 & 28/12/2015 10:51:08 - 10:58:30 & Turn on P-102. Pipe bursts. & Yes & Yes & Yes & The PLC .10 and .20 have highest scores. .60 raises often alarm when .10 is disturbed. This is expected because system is closed. Attention edges show correlation between .10, .60 and .20. \\ \hline
2015 & 3 & 28/12/2015 11:22:00 - 11:28:22 & Increase by 1 mm every second. Tank Underflow; Damage P101 & Yes & No & Partially & Attack Detected in .60-,.40- and .30. The attack was to .10. .60 and .10 are connected in closed loop, and therefore, any attacks through .10 might resonate in .60. \\ \hline
2015 & 7 & 28/12/2015 12:08:25 - 12:15:33 & Water level in-creased above HH. Stop of inflow; Tank underflow; Damage P301. & Yes & Yes & No & Attack detected in .30 and .40. The attention edges remain inconclusive. Most of the edges originate from .10 which was not part of the attack. \\ \hline
2015 & 8 & 28/12/2015 13:10:10 - 13:26:13 & Set value of DPIT as >40kpa. Backwash process is started again and again; Normal operation stops; Decrease in water level of tank 401. Increase in water level of tank 301. & Yes & Yes & Yes & Attack detected in .30, .40, and .60, as expected. Highest attention weights show correct origins. \\ \hline
2015 & 11 & 28/12/2015 14:19:00 - 14:28:20 & Set value of FIT-401 as 0. UV shutdown; P-501 turns off. & Yes & Yes & Partially & Correct source. The PLC:s around .40 raise alarms. One of three highest attention edges show relation between .40 and .60. The rest point to .10, which is considered incorrect. \\ \hline
2015 & 17 & 29/12/2015 14:38:12 - 14:50:08 & Do not let MV‐303 open. Halt of stage 3 because change in the backwash process. & Yes & Yes & Partially & Alarm in .60 and .30. Known behaviour for backwash to affect the 60. Part of the edges point to .20, which is considered incorrect.  \\ \hline
2015 & 19 & 29/12/2015 18:12:30 & Set value of AIT-504 to 16 uS/cm. RO shut down sequence starts after 30 minutes. Water should go to drain. & Yes & No & No & Alarm is not raised in .50. The .60 and .20 are the biggest anomaly sources. No evidence of the highest attention weights either.  \\ \hline
2015 & 21 & 9/12/2015 18:30:00 -- 18:42:00 & Keep MV-101 on countinuosly; Value of LIT-101 set as 700 mm. Tank overflow. & Yes & Yes & Yes & Alarm raised in .10, .60 and .20, as expected. .10 is clearly largest source. The attention edges from .10 support the detected anomalies, pointing to them. \\ \hline
2015 & 22 & 29/12/2015 22:55:18 - 23:03:00 & Stop UV401; Value of AIT502 set as 150; Force P501 to remain on. Possible damage to RO. & Yes & Yes & No & Alarm raised just after the end of attack, in .40, but not in .50. Largest edges between .60 and .10. Inconclusive.  \\ \hline
2015 & 23 & 30/12/2015 01:42:34 -- 01:54:10 & Value of PIT301 set to >0.4 bar; Keep MV302 open; Keep P602 closed. System freeze. & Yes & Yes & No & Correct detection of anomaly in .60 and .30. However, the attention edges are pointing to .10. The model fails to detect the causality between anomaly devices. \\ \hline
2015 & 26 & 30/12/2015 17:04:56 - 17:29:00 & P-101 is turned on continuosly; Set value of LIT-301 as 801 mm. Tank 101 underflow; Tank 301 overflow. & Yes & No & No & Alarm contribution mostly from .60. Attention weights are uniformly distributed, and therefore, analysis is inconclusive.  \\ \hline
2015 & 27 & 31/12/2015 01:20:20 & Keep P-302 on contineoulsy; Value of  LIT401 set as 600 mm till 1:26:01. Tank overflow. & Yes & No & No & No evidence, attention weights uniformly distributed accross PLCs.   \\ \hline
2015 & 28 & 31/12/2015 01:17:08 - 01:45:18 & Close P‐302. Stop inflow of tank T-401. & Yes & Yes & Partially & A known cascade failure, We detect a uniform alarm in all PLC:s, mostly in .60.  \\ \hline
2015 & 29 & 31/12/2015 15:32:00 - 15:34:00 & Turn on P201; Turn on P203; Turn on P205. Wastage of chemicals. & Yes & No & No & Immediate alert at the beginning. Attack not detected correctly in PLC .20. Attention edges do not reveal strong weights from .20, but rather, the edges are uniformly distributed between devices.   \\ \hline
2015 & 30 & 31/12/2015 15:47:40 - 16:07:10 & Turn P-101 on continuously; Turn MV-101 on continuously; Set value of LIT-101 as 700 mm; P-102 started itself because LIT301 level became low. Tank 101 underflow; Tank 301 overflow. & Yes & Yes & Partially & When attacked to .10, .60 is alerted. In this case, no other alarms are raised. Partial evidence of correct edges between devices, for example, between .10 and .30. \\ \hline
2015 & 32 & 1/1/2016 10:36:00 - 10:46:00 & Set LIT301 to above HH. Tank underflow; Damage P302. & Yes & Yes & Yes & Attack instantly raised in the beginning of the attack. The contribution to anomaly is correctly assigned to .30. Attention edges to/from SCADA-point. Not a typical pattern.   \\ \hline
2015 & 33 & 1/1/2016 14:21:12 - 14:28:35 & Set LIT‐101 to above H. Tank underflow; Damage P-101. & Yes & Yes & Yes & Alarm in .10 and .60. Edges are strong from/to .10 and the pattern is very plausible.   \\ \hline
2015 & 35 & 1/1/2016 17:21:40 & Turn P-101 off; Keep P-102 off. Stops outflow. & Yes & No & Partially & Surprisingly, the anomaly is not detected in .10, but clearly largest error in .30 and .20 Similar results were acquired in physical-level analysis. We find some evidence of correct attention edges, originating from .10.   \\ \hline
2015 & 36 & 1/1/2016 22:16:01 - 22:25:00 & Set LIT‐101 to less than LL. Tank overflow. & Yes & No & Yes & Again, for some reason, .30 contributes most when .10 is attacked. Again, we find strong evidence of correct attetion edges.   \\ \hline
2015 & 37 & 2/1/2016 11:17:02 - 11:24:50 & Close P501; Set value of FIT502 to 1.29 at 11:18:36. Reduced output. & Yes & Yes & No & Attack detected at the end of attack. PLC .50 detected, although attacks highest contribution is coming from PLC.60. Attention edges are uniformly distributed, and inference is not possible at this case.   \\ \hline
2015 & 39 & 2/1/2016 11:43:48 - 11:50:28 & Set value of FIT-401 as 0.5; Set value of AIT-502 as 140 mV. UV will shut down and water will go to RO. & Yes & Yes & Partially & .40 and .60 an .30 alarmed. .40 alarmed, a controller of UV. The attention edge pattern is a bit vague, as .60 and .30 have the highest edge weights.    \\ \hline
2015 & 41 & 2/1/2016 13:13:02 -- 13:40:56 & Decrease value by 0.5 mm per second. Tank overflow. & Yes & Yes & Yes &  .30 recognised. Also highest edge weights originate from .30 to .40 and .60, which are considered correct causality. \\ \hline
 \hline\hline
\end{longtable}
}
\end{landscape}

\bibliographystyle{unsrt}
\bibliography{sample}

\end{document}